\documentclass[]{elsart}

\usepackage{mathrsfs}
\usepackage{amsfonts}

\usepackage{amsmath}
\usepackage{algorithm}
\usepackage{booktabs}
\usepackage{array}
\usepackage{threeparttable}
\usepackage{multirow}
\usepackage{algpseudocode}
\usepackage{helvet}
\usepackage{courier}
\usepackage{tikz}
\usetikzlibrary{bayesnet,positioning}
\usepackage{subfigure}

\usepackage{graphicx}
\usepackage{endfloat}   

\usepackage{subfigure,url}

\usepackage{amssymb,bm}

\begin{document}

\begin{frontmatter}



\title{Online Anomaly Detection with Sparse Gaussian Processes}

\author{Jingjing Fei}
\author{, Shiliang Sun\corauthref{cor1}}
\corauth[cor1]{Corresponding author.
Tel.: +86-21-62233507;\\{\it E-mail address:}
{\rm shiliangsun@gmail.com, slsun@cs.ecnu.edu.cn}}
\address{Department of Computer Science and Technology,
East China Normal University, 3663 North Zhongshan Road, Shanghai 200241, P. R. China}

\begin{abstract}
Online anomaly detection of time-series data is an important and challenging task in machine learning. Gaussian processes (GPs) are powerful and flexible models for modeling time-series data. However, the high time complexity of GPs limits their applications in online anomaly detection. Attributed to some internal or external changes, concept drift usually occurs in time-series data, where the characteristics of data and meanings of abnormal behaviors alter over time. Online anomaly detection methods should have the ability to adapt to concept drift.
Motivated by the above facts, this paper proposes the method of sparse Gaussian processes with Q-function (SGP-Q). The SGP-Q employs sparse Gaussian processes (SGPs) whose time complexity is lower than that of GPs, thus significantly speeding up online anomaly detection. By using Q-function properly, the SGP-Q can adapt to concept drift well. Moreover, the SGP-Q makes use of few abnormal data in the training data by its strategy of updating training data, resulting in more accurate sparse Gaussian process regression models and better anomaly detection results. We evaluate the SGP-Q on various artificial and real-world datasets. Experimental results validate the effectiveness of the SGP-Q.
\end{abstract}


\end{frontmatter}

\section{Introduction}
The arrival of the Internet of Things (IoT) \cite{da2014internet} has inspired companies to install more sensors on their machines. These sensors can produce vast amounts of data which change over time and are called time-series data. The time-series data in our life are increasing rapidly, and it is of considerable significance and challenge to effectively process and monitor the time-series data.

A typical application scenario of time-series data is to monitor time-series data and send an alarm message to the operation engineers when an abnormal behavior occurs in the time-series data. This kind of application is called anomaly detection of time-series data. An abnormal behavior means that the current behavior is largely different from the normal behaviors in the past, and the current behavior is very rare \cite{chandola2008comparative,ahmad2017unsupervised}. In the real industrial environment, the time-series data are generated at all times, and only the previous and current data can be known, while the data after the current time are unseen. Therefore, online learning is essential for the anomaly detection of time-series data. Online anomaly detection determines whether the current behavior is abnormal or not according to the information of the current and previous data.

Online anomaly detection is significant because abnormal data can often convey essential and critical information in an extensive range of applications \cite{chandola2009anomaly,gupta2014outlier,habeeb2018real,ramotsoela2018survey,salehi2018survey,fernandes2019comprehensive,noble2018real,chalapathy2019deep,zhao2018online,sonmez2018anomaly,farshchi2018metric}. For example, when abnormal traffic occurs in a computer network, a hacker may be using the attacked computer to send sensitive data to the target computer \cite{kumar2005parallel}. When the MRI image is abnormal, it is likely to be due to the existence of some malignant tumor \cite{spence2001detection}. If abnormal data come from the sensors of the aircraft, it indicates that some parts of the aircraft may have malfunctions that need to be repaired in time \cite{fujimaki2005approach}.

Gaussian processes (GPs) are powerful and flexible tools for modeling time-series data \cite{GPML}. However, there are few methods of online anomaly detection based on GPs. There are two main reasons. Firstly, the time complexity of GPs is the cube of the number of training data. Training the GP model is time-consuming. Secondly, GPs are Bayesian nonparametric probabilistic models that are unfamiliar to industrial engineers. In recent years, many research work on sparse Gaussian processes (SGPs) have been proposed to reduce the high time complexity of GPs \cite{williams2001,herbrich2003fast,quinonero2005unifying,snelson2006sparse,titsias2009variational,hensman2013gaussian}. These research work can be divided into four categories. The first kind of method uses the {N}ystr{\"o}m method to approximate the covariance matrix \cite{williams2001,sun2015review}. The second kind of method employs a subset of data and selects informative points from the subset of data \cite{herbrich2003fast,liu2017sparse}. The third kind of method uses pseudo-inputs for the sparse approximation of GPs. The fourth kind of method employs variational inference and introduces inducing variables \cite{titsias2009variational,hensman2013gaussian}. SGPs greatly reduce the time complexity of GPs, which allows SGPs to be widely used on various types of data \cite{gal2014distributed,deisenroth2015distributed}. Inspired by the above facts, this paper employs SGPs with variational inducing variables for online anomaly detection.

The existing online anomaly detection methods based on GPs, including the Gaussian process regression with anomaly detection strategy (GPR-AD), Gaussian process regression with anomaly detection and mitigation strategy (GPR-ADAM) and Gaussian process regression
with the improved anomaly detection and mitigation strategy (GPR-IADAM) \cite{pengyu2017gpriadam}, cannot address concept drift in data properly. Concept drift means that data changes over time and the characteristics of new data are different from those of old data \cite{ahmad2017unsupervised,tsymbal2004problem,gama2014survey}. For example, sensor data collected by machines often changes on account of restarting machines or updating configurations. When the characteristics of data change, the definition of abnormal behaviors of data will also change. If anomaly detection methods cannot update their definition of abnormal behaviors in time, they cannot accurately detect anomalies in new data. Therefore, it is essential that online anomaly detection methods have the ability to adapt to the concept drift \cite{ma2018robust,khamassi2018discussion,almeida2018adapting,wang2018systematic,demvsar2018detecting,goldenberg2018survey,escovedo2018detecta}.

This paper proposes the method of sparse Gaussian processes with Q-function (SGP-Q). The SGP-Q uses the Q-function \cite{ahmad2017unsupervised,karagiannidis2007improved} to adapt to the concept drift in the data well. Specifically, the SGP-Q employs Q-function to measure the abnormal degree of the current data point relative to that of previous data. When the concept drift occurs in the data, the new data that initially change will be considered as abnormal data. However, when the `abnormal' behaviors persist for a while, the SGP-Q considers that the abnormal degree of the current data point relative to that of previous data is low. Then the SGP-Q adds the current data point and its time into training data to update the sparse Gaussian process regression (SGPR) model. The SGPR model can relearn the characteristics of new data to redefine the meanings of abnormal behaviors. Therefore, the SGP-Q can adapt to concept drift well. In the experiment, the proposed SGP-Q is compared with the online anomaly detection methods based on GPs, and experimental results validate the effectiveness of the proposed SGP-Q.

The contributions of this paper are listed as follows. Firstly, the proposed SGP-Q employs Q-function to measure the abnormal degree of the current data point relative to that of previous data, which can adapt to concept drift well. Secondly, SGPs with variational inducing variables are used to model time-series data in the SGP-Q, whose time complexity is much lower than that of GPs, thus speeding up online anomaly detection. Thirdly, the SGP-Q updates training data by the strategy based on likelihood and Q-function, which can make use of few abnormal data in the training data, thus making the trained SGPR model more accurate and anomaly detection results better.

The remainder of this paper is organized as follows. Section 2 describes the related work, including the introduction of GPs and several online anomaly detection methods based on GPs. Section 3 briefly reviews SGPs and introduces the proposed method SGP-Q in detail. Section 4 illustrates experiments, including the introduction of datasets, experimental setting, experimental results, and summary. Section 5 concludes the work of this paper.

\section{Related Work}
In this section, we will introduce GPs and several anomaly detection methods of time-series data based on GPs.
\subsection{Gaussian Processes}
GPs are powerful and flexible Bayesian nonparametric probabilistic models, which are mainly applied to regression and classification tasks. The GP regression model predicts continuous values, and the GP classification model predicts discrete values.

Assume that the size of training dataset $D$ is $N$, that is, $D=\{(\mathbf x_i,y_i)\}_{i=1}^{N}$, where $\mathbf x_i \in \mathbb R^Q$ and $y_i \in \mathbb R^+$ represent the input and output of the $i$th data point, respectively. Denote all inputs as $X$, and denote all outputs as $\mathbf y$.

GPs can be seen as Gaussian distributions on real-valued functions. The Gaussian distribution is uniquely determined by its mean and covariance matrix, and the GP is uniquely specified by its mean and covariance function similarly \cite{GPML}. A noiseless GP $f(\mathbf x)$ can be expressed as follows,
\begin{equation}\label{noiselessGP}
  f(\mathbf x) \sim \mathcal{GP}(m(\mathbf x), k(\mathbf x, \mathbf x')),
\end{equation}
where $m(\mathbf x)$ is the mean function of the GP and $k(\mathbf x, \mathbf x')$ is the covariance function of the GP, which is also known as kernel function. The selection of kernel function plays an important roles in the GP model. Common kernel functions include the radial basis function (RBF) kernel function, periodic kernel function and linear kernel function. The periodic kernel function is capable of modeling periodicity in data. These kernel functions are computed as follows,
\begin{equation}
\begin{aligned}
k_{(rbf)}(\mathbf x, \mathbf x') &= \sigma_{rbf}^2\exp\left[-\frac{|\mathbf x-\mathbf x'|^2}{2l^2}\right],\\
k_{(periodic)}(\mathbf x, \mathbf x') &= \sigma_{periodic}^2 \exp \left[  - \frac{1}{2} \sum_{i=1}^{Q}\left( \frac{\sin(\frac{\pi}{T} (x_i - x'_i) )}{l_i} \right)^2 \right],\\
k_{(linear)}(\mathbf x, \mathbf x') &= \sum_{i=1}^{Q} \sigma^2_{linear(i)} x_ix'_i,
\end{aligned}
\end{equation}
where $\sigma_{rbf}^2$, $\sigma_{periodic}^2$ and $\sigma_{linear}^2$ are the variances, $l$ is the length-scale and $T$ is the periodic parameter.

The mean and covariance functions are calculated as follows,
\begin{equation}
\begin{aligned}
\centering
  m(\mathbf x) &= \mathbb{E}[f(\mathbf x)],\\
  k(\mathbf x, \mathbf x') &= \mathbb{E}[(f(\mathbf x)-m(\mathbf x))(f(\mathbf x')-m(\mathbf x'))].
\end{aligned}
\end{equation}

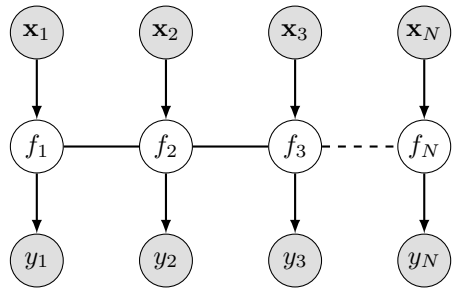
\begin{figure}[t]
\centering
\begin{tikzpicture}[scale = 0.2]
        \node[obs,minimum size = 20pt] (y1) {${y_1}$} ; %
        \node[obs,right=of y1] (y2) {${y}_2$} ; %
        \node[obs,right=of y2] (y3) {${y}_3$} ; %
        \node[obs,right=of y3] (yN) {${y}_N$} ; %
        \node[latent, above=of y1,yshift=-0.2cm] (f1) {$f_1$} ; %
         \node[latent,right=of f1] (f2) {${f}_2$} ; %
         \node[latent,right=of f2] (f3) {${f}_3$} ; %
        \node[latent,right=of f3] (fN) {${f}_N$} ; %

        \tikzstyle{connect}=[-latex, thick]
        \tikzstyle{connectf}=[-, thick]
        \tikzstyle{connectfdashed}=[-, thick, dashed]

        \node[obs, above=of f1,yshift=-0.2cm] (x1) {$\mathbf{x}_1$} ; %
        \node[obs, above=of f2,yshift=-0.2cm] (x2) {$\mathbf{x}_2$} ; %
         \node[obs, above=of f3,yshift=-0.2cm] (x3) {$\mathbf{x}_3$} ; %
        \node[obs, above=of fN,yshift=-0.2cm] (xN) {$\mathbf{x}_N$} ; %
        \path (x1) edge [connect] (f1);
        \path (x2) edge [connect] (f2);
        \path (x3) edge [connect] (f3);
        \path (xN) edge [connect] (fN);
        \path (f1) edge [connect] (y1);
        \path (f2) edge [connect] (y2);
        \path (f3) edge [connect] (y3);
        \path (fN) edge [connect] (yN);
        \path (f1) edge [connectf] (f2);
        \path (f2) edge [connectf] (f3);
        \path (f3) edge [connectfdashed] (fN);
\end{tikzpicture}
\caption{The graphical model of a GP with $N$ training points.}
\label{figure:GP}
\end{figure}

In the noiseless situation, the output $y$ is the GP $f(\mathbf x)$. However, in general, the observed output $y$ is not the GP $f(\mathbf x)$, but with some noise, e.g., $y = f(\mathbf x) + \epsilon$, where $\epsilon \sim N(0,\sigma^2)$ is the additive noise. Figure \ref{figure:GP} shows the graphical model of a GP.

Given the model assumption of the GP, the Gaussian likelihood is $p(\mathbf y|\mathbf f) = \mathcal{N}(\mathbf f, \sigma^2I)$. After integrating the latent variable $\mathbf f$, we can get the marginal likelihood distribution $p(\mathbf y|X) = \mathbf N(0,K_{NN}+\sigma^2I)$ where $K_{NN} = K(X,X)$ is an $N\times N$ covariance matrix. The GP model is learned by maximizing the marginal likelihood. The posterior of the latent variable $\mathbf f$ is as follows,
\begin{equation}
  p(\mathbf f|\mathbf y) = \mathcal N(\bm \mu, \Sigma),
\end{equation}
where the mean $\bm \mu = K_{NN}(K_{NN}+\sigma^2I)^{-1}\mathbf y$ and the covariance matrix $\Sigma = K_{NN}-K_{NN}(K_{NN}+\sigma^2I)^{-1}K_{NN}$.

Given a new input $\mathbf x^*$, the prediction distribution is still a Gaussian distribution,
\begin{equation}\label{gp:prediction}
  p(y^*|X,\mathbf y,\mathbf x^*) = N(\mu^*,\Sigma_y^*),
\end{equation}
where $\mu^* = {k(\mathbf x^*,X)}[K_{NN}+\sigma^2I]^{-1}\mathbf y$ and $\Sigma_y^* = k(\mathbf x^*, \mathbf x^*) - k(\mathbf x^*,X)(K_{NN}+\sigma^2I)^{-1}k(X, \mathbf x^*)+\sigma^2$. The time complexity of the GP is $O(N^3)$, where $N$ is the number of training data.

\subsection{Gaussian Processes for Anomaly Detection in Time-Series Data}
In this section, we mainly introduce several existing anomaly detection methods of time-series data based on GPs, including the Gaussian process regression (GPR) with the anomaly detection strategy (GPR-AD), the GPR with the anomaly detection and mitigation strategy (GPR-ADAM) and the GPR with the improved anomaly detection and mitigation strategy (GPR-IADAM) \cite{pengyu2017gpriadam}. These methods belong to anomaly detection methods based on prediction, which detect anomalies by judging whether the data fall into the prediction interval.

Firstly, we focus on the state-of-the-art anomaly detection method of time-series data, that is, the GPR-IADAM.
For the task of anomaly detection in time-series data, the input is time. Assuming the current time is $m$, the training data is composed of $q$ data points that are closest to the current moment, i.e., $D_T=\{t_i,y_i\}_{i=m-q+1}^m$. The GPR model can be obtained by maximizing the marginal likelihood. The time $t_{m+1}$ at the next moment is taken as the test input, and the prediction mean $\mu_{m+1}$ and variance $\sigma^2_{m+1}$ can be obtained according to the Equation (\ref{gp:prediction}). The 95 percent confidence interval for the Gaussian distribution is $[\mu_{m+1} - 1.96\sigma_{m+1}, \mu_{m+1} + 1.96\sigma_{m+1}]$.

If the test data point $y_{m+1}$ is within the 95 percent confidence interval, it is considered as a normal data point, and the data point $y_{m+1}$ and its time $t_{m+1}$ are added into the sliding window $D_T$. If the test data point $y_{m+1}$ is not in the 95 percent confidence interval, it is marked as an abnormal data point. The value of $\beta(y_{m+1})$ needs to be calculated according to the Equation (\ref{beta}).
\begin{equation}\label{beta}
  \beta(y_{m+1}) = P(z<1.96-\frac{|\mu_{m+1}-y_{m+1}|}{\sigma_{m+1}}),
\end{equation}
where $z$ obeys the standard Gaussian distribution.
The value of $\beta(y_{m+1})$ is used to measure the deviation between the data point $y_{m+1}$ and prediction mean $\mu_{m+1}$. The smaller the value of $\beta$ is, the greater the deviation between the data point $y_{m+1}$ and prediction mean $\mu_{m+1}$ is. The larger the value of $\beta$ is, the smaller the deviation between the data point $y_{m+1}$ and prediction mean $\mu_{m+1}$ is. Comparing the value of $\beta$ and $\beta_{max}$, if $\beta$ is less than or equal to $\beta_{max}$, the prediction mean $\mu_{m+1}$ and its time $t_{m+1}$ will be added to the sliding window $D_T$. If $\beta$ is greater than $\beta_{max}$, the data point $y_{m+1}$ and its time $t_{m+1}$ will be added to the sliding window $D_T$. Here $\beta_{max}$ is an artificially specified threshold, and the GPR-IADAM sets the value of $\beta_{max}$ to 0.05.

After adding the new data point and its time in the sliding window $D_T$, the earliest data point and its time are removed from the $D_T$. The data in the new sliding window $D_T$ are then employed to update the GPR model. Algorithm \ref{alg:GPR-IADAM} shows the GPR-IADAM method.

\begin{algorithm}[h]
\caption{GPR-IADAM}
\label{alg:GPR-IADAM}
\begin{algorithmic}[1]
\Require current time $m$, $D_T = \{t_i,y_i\}_{i=m-q+1}^{m}$, the size of sliding window $q$ and threshold $\beta_{max}$
\Ensure select an appropriate covariance function (the mean function is generally set to 0), and initialize parameters of the covariance function
\Repeat
\State train the GPR model using data in the sliding window $D_T$
\State predict mean $\mu_{m+1}$ and variance $\sigma^2_{m+1}$ at time $t_{m+1}$ by Equation (\ref{gp:prediction})
\State compute 95\% confidence interval $[\mu_{m+1}-1.96\sigma_{m+1},\mu_{m+1} + 1.96\sigma_{m+1}]$
\If{data point $y_{m+1}$ in 95\% confidence interval}
\State $y_{m+1}$ is a normal data point
\State add $\{t_{m+1},y_{m+1}\}$ into $D_T$
\Else
\State $y_{m+1}$ is an abnormal data point
\State compute the value of $\beta(y_{m+1})$ by Equation (\ref{beta})
\If {$\beta(y_{m+1}) \le \beta_{max}$}
\State add $\{t_{m+1},\mu_{m+1}\}$ into $D_T$
\Else
\State add $\{t_{m+1}, y_{m+1}\}$ into $D_T$
\EndIf
\EndIf
\State remove the earliest data point and its time from $D_T$
\State $m=m+1$
\Until{all test data points have been detected}
\label{code:recentEnd}
\end{algorithmic}
\end{algorithm}

Secondly, we will introduce the GPR-AD. The difference between the GPR-AD and GPR-IADAM is the strategy of updating data in the sliding window $D_T$. The GPR-AD updates the sliding window $D_T$ using the anomaly detection (AD) strategy. The AD strategy adds the data point $y_{m+1}$ and its time $t_{m+1}$ to the sliding window $D_T$ regardless of whether the data point is abnormal or not, and the earliest data point and its time are removed from the $D_T$. Algorithm \ref{alg:GPR-AD} shows the GPR-AD method.

\begin{algorithm}[h]
\caption{GPR-AD}
\label{alg:GPR-AD}
\begin{algorithmic}[1]
\Require current time $m$, $D_T = \{t_i,y_i\}_{i=m-q+1}^{m}$ and the size of sliding window $q$
\Ensure select an appropriate covariance function (the mean function is generally set to 0), and initialize parameters of the covariance functions
\Repeat
\State train the GPR model using data in the sliding window $D_T$
\State predict mean $\mu_{m+1}$ and variance $\sigma_{m+1}$ at time $t_{m+1}$ by Equation (\ref{gp:prediction})
\State compute 95\% confidence interval $[\mu_{m+1}-1.96\sigma_{m+1},\mu_{m+1} + 1.96\sigma_{m+1}]$
\If{data point $y_{m+1}$ in 95\% confidence interval}
\State $y_{m+1}$ is a normal data point
\Else
\State $y_{m+1}$ is an abnormal data point
\EndIf
\State add $\{t_{m+1},y_{m+1}\}$ into $D_T$
\State remove the earliest data point and its time from $D_T$
\State $m=m+1$
\Until{all test data points have been detected}
\label{code:recentEnd}
\end{algorithmic}
\end{algorithm}

\begin{algorithm}[h]
\caption{GPR-ADAM}
\label{alg:GPR-ADAM}
\begin{algorithmic}[1]
\Require current time $m$, $D_T = \{t_i,y_i\}_{i=m-q+1}^{m}$ and the size of sliding window $q$
\Ensure select an appropriate covariance function (the mean function is generally set to 0), and initialize parameters of the covariance function
\Repeat
\State train the GPR model using data in sliding window $D_T$
\State predict mean $\mu_{m+1}$ and variance $\sigma_{m+1}$ at time $t_{m+1}$ by Equation (\ref{gp:prediction})
\State compute 95\% confidence interval $[\mu_{m+1}-1.96\sigma_{m+1},\mu_{m+1} + 1.96\sigma_{m+1}]$
\If{data point $y_{m+1}$ in 95\% confidence interval}
\State $y_{m+1}$ is a normal data point
\State add $\{t_{m+1},y_{m+1}\}$ into $D_T$
\Else
\State $y_{m+1}$ is an abnormal data point
\State add $\{t_{m+1},\mu_{m+1}\}$ into $D_T$
\EndIf
\State remove the earliest data point and its time from $D_T$
\State $m=m+1$.
\Until{all test data points have been detected}
\label{code:recentEnd}
\end{algorithmic}
\end{algorithm}

Last but not least, we briefly describe the GPR-ADAM. The only difference between the GPR-IADAM and GPR-ADAM is also the strategy of updating data in the sliding window $D_T$. GPR-ADAM updates the sliding window $D_T$ using the anomaly detection and mitigation (ADAM) strategy. In the ADAM strategy, the data point $y_{m+1}$ and its time $t_{m+1}$ are added to the sliding window $D_T$ when the data point is normal. The prediction mean $\mu_{m+1}$ and its time $t_{m+1}$ are added to the sliding window $D_T$ when the data point is abnormal. The earliest data point and its time are removed from the $D_T$. Algorithm \ref{alg:GPR-ADAM} shows the GPR-ADAM method.

\section{Sparse Gaussian Processes with Q-Function}
In this section, we first review the sparse Gaussian processes (SGPs) and then introduce the proposed method of sparse Gaussian processes with Q-function (SGP-Q).
\subsection{Sparse Gaussian Processes}
The GP is a powerful and flexible model, but its high time complexity $O(N^3)$ limits its application scenarios. In order to reduce the high time complexity of the GP, different sparse approximation methods are proposed \cite{williams2001,herbrich2003fast,quinonero2005unifying,snelson2006sparse,titsias2009variational,hensman2013gaussian}.
Here we mainly review the sparse approximation method using variational inducing variables \cite{quinonero2005unifying,titsias2009variational}.

In the approximation methods based on inducing variables, the active set is not a subset of data selected from training data but is the inducing input $Z$ obtained through optimization. The latent variable corresponding to the inducing input is $\mathbf u=f(Z)$. SGPs with inducing variables usually have three forms of approximation, that is, deterministic training conditional (DTC) approximation, fully independent training conditional (FITC) approximation and partially independent training conditional (PITC) approximation \cite{quinonero2005unifying}. The key difference between the three approximation methods is that they assume different conditional distributions $p(\mathbf f|\mathbf u, Z, X)$.

The DTC approximation refers to the fact that the value of the latent variable $\mathbf f$ is deterministic when the inducing variable $\mathbf u$ is known,
\begin{equation}\label{DTCLikelihood}
  p(\mathbf f|\mathbf u,Z,X) = \mathcal{N}(\mathbf f|K_{NM}K_{MM}^{-1}\mathbf u, \mathbf 0),
\end{equation}
where $M$ is the number of data points in the inducing input, $K_{NM} = K(X,Z)$, $K_{MM} = K(Z,Z)$ and $K_{MN} = K(Z,X)$.

The FITC approximation means that the latent variable $\mathbf f$ is fully independent when the inducing variable $\mathbf u$ is known,
\begin{equation}\label{FITCLikelihood}
  p(\mathbf f|\mathbf u,Z,X) = \mathcal{N}(\mathbf f|K_{NM}K_{MM}^{-1}\mathbf u, diag[K_{NN}-K_{NN}K_{MM}^{-1}K_{MN}]).
\end{equation}

The PITC approximation signifies that the latent variable $\mathbf f$ is partially independent when the inducing variable $\mathbf u$ is known,
\begin{equation}\label{FITCLikelihood}
  p(\mathbf f|\mathbf u,Z,X) = \mathcal{N}(\mathbf f|K_{NM}K_{MM}^{-1}\mathbf u, blockdiag[K_{NN}-K_{NN}K_{MM}^{-1}K_{MN}]).
\end{equation}

Under the above three assumptions, the marginal likelihood of the SGP is a function of the inducing input and hyperparameters of the kernel functions. The inducing input and hyperparameters can be obtained by maximizing the marginal likelihood, which may lead to over-fitting.

SGPs with variational inducing variables assume the augmented variational posterior $q(\mathbf f,\mathbf u)$ to approximate the augmented true posterior $p(\mathbf f, \mathbf u|\mathbf y)$, and minimize the Kullback-Leibler (KL) divergence between the variational posterior and true posterior $KL(q(\mathbf f,\mathbf u)||p(\mathbf f, \mathbf u|\mathbf y))$ \cite{titsias2009variational}. Minimizing the KL divergence $KL(q(\mathbf f,\mathbf u)||p(\mathbf f, \mathbf u|\mathbf y))$ is equivalent to maximizing the variational lower bound of the marginal likelihood. The variational lower bound of the SGP is as follows,
\begin{equation}\label{SGPELBO}
  \mathcal{L} = \log\left[N(\mathbf y|\mathbf 0, \sigma^2I+K_{NM}K_{MM}^{-1}K_{MN})\right] - \frac{1}{2\sigma^2}Tr\left[K_{NN}-K_{NM}K_{MM}^{-1}K_{MN}\right].
\end{equation}
Here the inducing input $Z$ is seen as the variational parameter obtained by optimizing the variational lower bound and has no impact on the marginal likelihood $p(\mathbf y)$ and posterior $p(\mathbf f|\mathbf y)$, which is beneficial to avoiding over-fitting to a certain extent.

Given a test input $\mathbf x_*$, the prediction distribution of the SGP with variational inducing variables is still a Gaussian distribution as follows \cite{titsias2009variational},
\begin{equation}\label{sgp:prediction}
  p(y^*|X,Z,\mathbf y,\mathbf x^*) = N(\mu^*,\Sigma_y^*),
\end{equation}
where $\mu^* = k(\mathbf x^*,Z)K_{MM}^{-1}\tilde{\bm \mu}$, $\Sigma_y^* = k(\mathbf x^*,\mathbf x^*) - k(\mathbf x^*,Z)K_{MM}^{-1}k(Z,\mathbf x_*) + k(\mathbf x^*,Z)B\\k(Z,\mathbf x_*)+\sigma^2$, $\tilde{\bm \mu} = \sigma^{-2}K_{MM}\Sigma K_{MN}\mathbf y$, $A = K_{MM}\Sigma K_{MM}$, $\Sigma = (K_{MM}+\sigma^{-2}K_{MN}K_{NM})^{-1}$, and $B = K_{MM}^{-1}AK_{MM}^{-1}$.
The time complexity of the SGP with variational inducing variables is $O(NM^2)$.

\subsection{Online Anomaly Detection with Sparse Gaussian Processes}
The existing online anomaly detection methods have some limitations. The GPR-AD adds a data point to the training data no matter whether the data point is abnormal or not. When lots of abnormal data are added into the training data, the prediction may lose its accuracy, thus leading to wrong anomaly detection results. The GPR-ADAM adds a data point into the training data when the data point is normal and adds the prediction mean of the GPR model to the training data when the data point is abnormal. The GPR-ADAM alleviates the negative impact of abnormal data on the training model. The GPR-IADAM is a compromise between the GPR-AD and GPR-ADAM. Specifically, the GPR-IADAM uses $\beta$ to measure the deviation between the data point and the prediction mean. If the prediction mean deviates significantly from the data point, the prediction mean is added to the training data. If the prediction mean deviates slightly from the data point, the data point will be added to the training data. The GPR-ADAM and GPR-IADAM decide to add the data point or prediction mean into the training data according to the abnormal degree of the current data point, which cannot solve the problem of concept drift in the data. Concept drift \cite{gama2014survey,pratama2016scaffolding,ahmad2017unsupervised} means that the characteristics of data and the mapping from input to output change over time. For example, when the computer's software is updated or its configuration is changed, data such as CPU utilization and the speed of reading or writing data in the disk will change. In this case, online anomaly detection methods should be able to adapt to the new data and redefine the meanings of abnormal behaviors.

To overcome the limitations of existing anomaly detection methods of time-series data based on GPs, this paper proposes the SGP-Q. On the one hand, the SGP-Q uses SGPs to model the relationship between time and observed data. The time complexity of SGPs is much lower than that of GPs, while their performance is similar. Therefore, SGPs are more suitable for the task of online anomaly detection. On the other hand, the SGP-Q method employs Q-function \cite{karagiannidis2007improved} to solve the problem of concept drift in time-series data.

Specifically, the SGP-Q uses the technique of sliding window, and the size of the sliding window is set to $q$. Suppose current time is $m$, then data in the sliding window can be represented as $D_T=\{t_i,y_i\}_{i=m-q+1}^m$. Given a test data point $\{t_{m+1},y_{m+1}\}$, the time $t_{m+1}$ is the input of the SGPR, and the outputs of the SGPR contain the prediction mean $\mu_{m+1}$ and variance $\sigma^2_{m+1}$.
Then the SGP-Q computes the likelihood of the data point $y_{m+1}$ by Equation (\ref{SGPQlikelihhood}).
\begin{equation}\label{SGPQlikelihhood}
  p(y_{m+1}) = \frac{1}{\sqrt{2\pi} \sigma_{m+1}}\exp\left[-\frac{(y_{m+1}-\mu_{m+1})^2}{2\sigma^2_{m+1}}\right].
\end{equation}

If the likelihood of the test data point $p(y_{m+1})$ is greater than or equal to the threshold $\epsilon_p$, the data point $y_{m+1}$ is normal, and the data point $y_{m+1}$ and its time $t_{m+1}$ are added to the sliding window $D_T$. If the likelihood of the test data point $p(y_{m+1})$ is less than the threshold $\epsilon_p$, the data point is abnormal. The SGP-Q lists a set of possible thresholds, and selects the threshold which performs best in the validation set according to the $F_1$ score as $\epsilon_p$. For an abnormal data point, the SGP-Q computes the Q-function based on the absolute error $e(y_{m+1}) = |y_{m+1}-\mu_{m+1}|$ and the Q-function based on the likelihood $p(y_{m+1})$. The computation of Q-function requires three steps \cite{ahmad2017unsupervised}.

Firstly, the SGP-Q employs two windows to accommodate the latest $W$ values of error and likelihood, respectively. The distributions of the error and likelihood are modeled as Gaussian distributions where the mean and variance are updated according to the latest values of error and likelihood as follows,
\begin{equation}\label{QfunMuSigmaW}
\begin{aligned}
\hat{\mu}(e(y_{m+1})) &= \frac{\sum_{i=0}^{i=W-1}e(y_{m+1-i})}{W},\\
\sigma^2(e(y_{m+1})) &= \frac{\sum_{i=0}^{i=W-1}[e(y_{m+1-i})-\hat{\mu}(e(y_{m+1}))]^2}{W-1},\\
\hat{\mu}(p(y_{m+1})) &= \frac{\sum_{i=0}^{i=W-1}p(y_{m+1-i})}{W},\\
\sigma^2(p(y_{m+1})) &= \frac{\sum_{i=0}^{i=W-1}[p(y_{m+1-i})-\hat{\mu}(p(y_{m+1}))]^2}{W-1}.\\
\end{aligned}
\end{equation}

Secondly, the SGP-Q computes the mean of the error and likelihood over the recent short period as follows,
\begin{equation}\label{QfunMuW1}
\begin{aligned}
  \tilde{\mu}(e(y_{m+1})) &= \frac{\sum_{i=0}^{i=W'-1}e(y_{m+1-i})}{W'},\\
  \tilde{\mu}(p(y_{m+1})) &= \frac{\sum_{i=0}^{i=W'-1}p(y_{m+1-i})}{W'},\\
\end{aligned}
\end{equation}
where $W'$ is the size of the window for the recent short period, and $W'\ll W$.

Thirdly, the Gaussian Q-function (denoted as Q-function) is of considerable significance in various fields, such as statistics and communication theory \cite{simon2007probability,zwillinger2014table}. The Q-function has no analytic solution, and several research work are proposed for approximate calculation of the Q-function \cite{karagiannidis2007improved,chiani2003new,borjesson1979simple}.
An approximation of the Q-function is the exponential approximation which is easy to calculate. The exponential approximation of the Q-function is $Q(x) \approx \frac{1}{12} \exp(-\frac{x^2}{2}) + \frac{1}{4}\exp(-\frac{2x^2}{3})$ \cite{chiani2003new}.
The SGP-Q modifies the original exponential approximation of the Q-function to make it more sensitive to slight input changes. Our modified Q-function is $Q(x) \approx \frac{1}{6} \exp(-\frac{x^2}{4}) + \frac{1}{2}\exp(-\frac{x^2}{3})$.
Our modified Q-function is an even function and the value of modified Q-function decreases as the absolute value of the input increases. Then the Q-function based on the absolute error (denoted as QE) and likelihood (denoted as QL) are defined as follows,
\begin{equation}\label{QFunErrPro}
  \begin{aligned}
  QE_{m+1} = Q\left(\frac{\tilde{\mu}(e(y_{m+1}))-\hat{\mu}(e(y_{m+1}))}{\sigma^2(e(y_{m+1}))}\right),\\
  QL_{m+1} = Q\left(\frac{\tilde{\mu}(p(y_{m+1}))-\hat{\mu}(p(y_{m+1}))}{\sigma^2(p(y_{m+1}))}\right).
  \end{aligned}
\end{equation}

Q-function can measure the abnormal degree of the current data point relative to that of previous data. The smaller the value of Q-function, the higher the abnormal degree of the current data point relative to that of previous data. The larger the value of Q-function, the lower the abnormal degree of the current data point relative to that of previous data.

For an abnormal data point, if at least one of the two conditions $QE_{m+1}<\epsilon_e$ and $QL_{m+1}<\epsilon_l$ is satisfied ($\epsilon_e=3e-1$ and $\epsilon_l=3e-1$), the SGP-Q adds the prediction mean $\mu_{m+1}$ and its time $t_{m+1}$ to the sliding window $D_T$. If neither of the two conditions is met, the data point $y_{m+1}$ and its time $t_{m+1}$ are added to the sliding window $D_T$. After adding the latest data point to the sliding window $D_T$, the earliest data point and its time are removed from the $D_T$. The new data in the sliding window $D_T$ are then employed to update the SGPR model. Algorithm \ref{alg:SGP-Q} shows the SGP-Q method.
\begin{algorithm}[h]
\caption{SGP-Q}
\label{alg:SGP-Q}
\begin{algorithmic}[1]
\Require current time $m$, $D_T = \{t_i,y_i\}_{i=m-q+1}^{m}$, the size of the window $q$, $W$, $W'$, threshold $\epsilon_e$, $\epsilon_l$, and the size of inducing points $M$
\Ensure select an appropriate covariance function (the mean function is generally set to 0), and initialize the parameters of the covariance function
\Repeat
\State train the SGPR model using data in sliding window $D_T$
\State predict mean $\mu_{m+1}$ and variance $\sigma^2_{m+1}$ at time $t_{m+1}$ by Equation (\ref{sgp:prediction})
\State compute the likelihood of data point $p(y_{m+1})$ by Equation (\ref{SGPQlikelihhood})
\State select the threshold which performs best in the validation set according to $F_1$ score as $\epsilon_p$
\If{$p(y_{m+1})>\epsilon_p$}
\State $y_{m+1}$ is a normal data point
\State add $\{t_{m+1},y_{m+1}\}$ into $D_T$
\Else
\State $y_{m+1}$ is an abnormal data point
\State compute the value of $QE_{m+1}$ and $QL_{m+1}$ by Equation (\ref{QFunErrPro})
\If {$QE_{m+1} \le \epsilon_e$ or $QL_{m+1} \le \epsilon_l$}
\State add $\{t_{m+1},\mu_{m+1}\}$ into $D_T$
\Else
\State add $\{t_{m+1}, y_{m+1}\}$ into $D_T$
\EndIf
\EndIf
\State remove the earliest data point and its time from $D_T$
\State $m=m+1$
\Until{all test data points have been detected}
\label{code:recentEnd}
\end{algorithmic}
\end{algorithm}

Unlike the GPR-ADAM and GPR-IADAM which uses the information of the current data point to measure the abnormal degree of the current data point, the SGP-Q employs information of previous and current data to measure the abnormal degree of the current data point relative to that of previous data.
The SGP-Q can address concept drift well. Specifically, when the concept drift occurs, data that start to change will be marked as abnormal data by the SGP-Q. However, when the `abnormal' behaviors continue for a while, the SGP-Q judges the abnormal degree of the current data point is low compared with that of previous data. Then the current data point $y_{m+1}$ and its time $t_{m+1}$ are added to the sliding window $D_T$ to train the SGPR model. Therefore, the SGP-Q can learn new characteristics of data and redefine the meanings of abnormal behaviors.

\section{Experiments}
In this section, we conduct experiments to validate the rationality and effectiveness of the proposed SGP-Q method.

\subsection{Data}
We conducted experiments on eight datasets from the Numenta Anomaly Benchmark (NAB) \cite{ahmad2017unsupervised}. Two of the eight datasets are the artificially generated datasets in NAB, denoted as `art\_daily\_jumpsup' and `art\_daily\_flatmiddle'. The number of data points on both artificially generated datasets is 4032. Four of the eight datasets are data of Amazon Web Services (AWS) server provided by the AmazonCloudwatch service. These four datasets of AWS server are denoted as 、ec2\_cpu\_utilization\_24ae8d', `ec2\_cpu\_utilization\_825cc2', `ec2\_cpu\_utilization\_ac20cd'  and `grok\_asg\_anomaly', and the number of data points on the four datasets is 4032, 4032, 4648 and 4621, respectively. The remaining two datasets are real-time traffic data of the Twin Cities Metro area in Minnesota which are offered by the Minnesota Department of Transportation. These two traffic datasets are denoted as `occupancy\_t4013' and `speed\_t4013', and the number of data points on the two traffic datasets is 2500 and 2495, respectively.

\subsection{Setting}
For the SGP-Q, GPR-AD, GPR-ADAM and GPR-IADAM, the timestamps in the NAB exist as strings and need to be preprocessed for calculation. We quantize the timestamps to the number of minutes between the current time and today's 00:00:00 multiplied by 0.01. The covariance function is set to the sum of the RBF kernel function and the linear kernel function. The number of points in the inducing input $Z$ is set to $M=100$. The size of the sliding window $D_T$ is set to $q=1000$. The size of windows in the Q-function are set to $W = 500$ and $W'=10$. For the GPR-IADAM, $\beta_{max} = 0.05$. When the model is trained for the first time, we set the number of iterations to 1000. When the newest data point is added to $D_T$ and the earliest data point is removed from $D_T$, we continue to optimize the model and set the number of iterations to 10. The data used for training the model for the first time is a piece of normal data at the beginning of each time-series data, where a small amount of abnormal data is allowed. The remaining data in each time-series data are used for testing. After the detection of each test data point, its real value or prediction mean will be added into the sliding window to update the model. A short time-series segment containing normal and abnormal data is taken from each time-series data, and then a tiny amount of noise is added to generate data in the validation set.

\subsection{Experimental Results}
We compare the proposed SGP-Q with the GPR-AD, GPR-ADAM, and GPR-IADAM. Since $F_1$ score can take into account the precision and recall, we use $F_1$ score as the measure of performance.
\begin{equation}\label{F1}
  F_1 = \frac{2\times precision\times recall}{precision+recall}.
\end{equation}

All experiments were repeated five times, and the average results were taken as the final results.
Table \ref{F1Results} shows the $F_1(\%)$ score of the four methods on eight datasets, and the best results are in bold. As shown in Table \ref{F1Results}, the proposed method SGP-Q performs best on all datasets, which demonstrates the effectiveness of the SGP-Q.
\begin{table}[H]
\centering
\begin{threeparttable}
\begin{tabular}{ | c | c |  c | c | c |}
\hline
Dataset & GPR-AD & GPR-ADAM & GPR-IADAM & SGP-Q \\ \hline
art\_daily\_jumpsup & 82.20$\pm$0.40 & 91.40$\pm$0.80 &  92.80$\pm$0.40 & \textbf{99.20$\pm$0.40}  \\ \hline
art\_daily\_flatmiddle & 74.20$\pm$0.45 & 93.20$\pm$0.45 &  93.20$\pm$0.45 & \textbf{96.40$\pm$1.09}  \\ \hline
ec2\_cpu\_utilization\_24ae8d & 98.90$\pm$0.74 & 98.70$\pm$0.45 &  98.70$\pm$0.45 & \textbf{99.00$\pm$0.82}  \\ \hline
ec2\_cpu\_utilization\_825cc2 & 93.60$\pm$0.55 & 44.40$\pm$0.89 &  86.40$\pm$0.55 & \textbf{99.60$\pm$0.55}  \\ \hline
ec2\_cpu\_utilization\_ac20cd & 95.40$\pm$0.55 & 29.40$\pm$0.55 &  30.20$\pm$0.45 & \textbf{96.20$\pm$1.17}  \\ \hline
grok\_asg\_anomaly & 85.20$\pm$0.84 & 52.00$\pm$0.00 & 55.00$\pm$0.00 & \textbf{90.00$\pm$1.41}  \\ \hline
occupancy\_t4013 & 80.00$\pm$0.00 & 80.40$\pm$0.90 & 80.00$\pm$0.00 & \textbf{83.00$\pm$1.22}  \\ \hline
speed\_t4013 & 81.40$\pm$0.89 & 77.60$\pm$0.54 & 81.60$\pm$0.55
& \textbf{84.00$\pm$0.00}  \\ \hline
\end{tabular}
\caption{The $F_1(\%)$ score of four methods on eight datasets. The best results are in bold.}
\label{F1Results}
\end{threeparttable}
\end{table}

In addition, we will visually show the results of anomaly detection. The figures below show the results of anomaly detection by four methods and true labels of test data on eight datasets. The blue crosses represent test data, and the red circles represent detected anomalies. When the label is equal to 1, the test data point is abnormal. When the label is equal to 0, the test data point is normal.

\begin{figure}[ht]
\centering
\subfigure[GPR-AD]
{\includegraphics[width=2.7in]{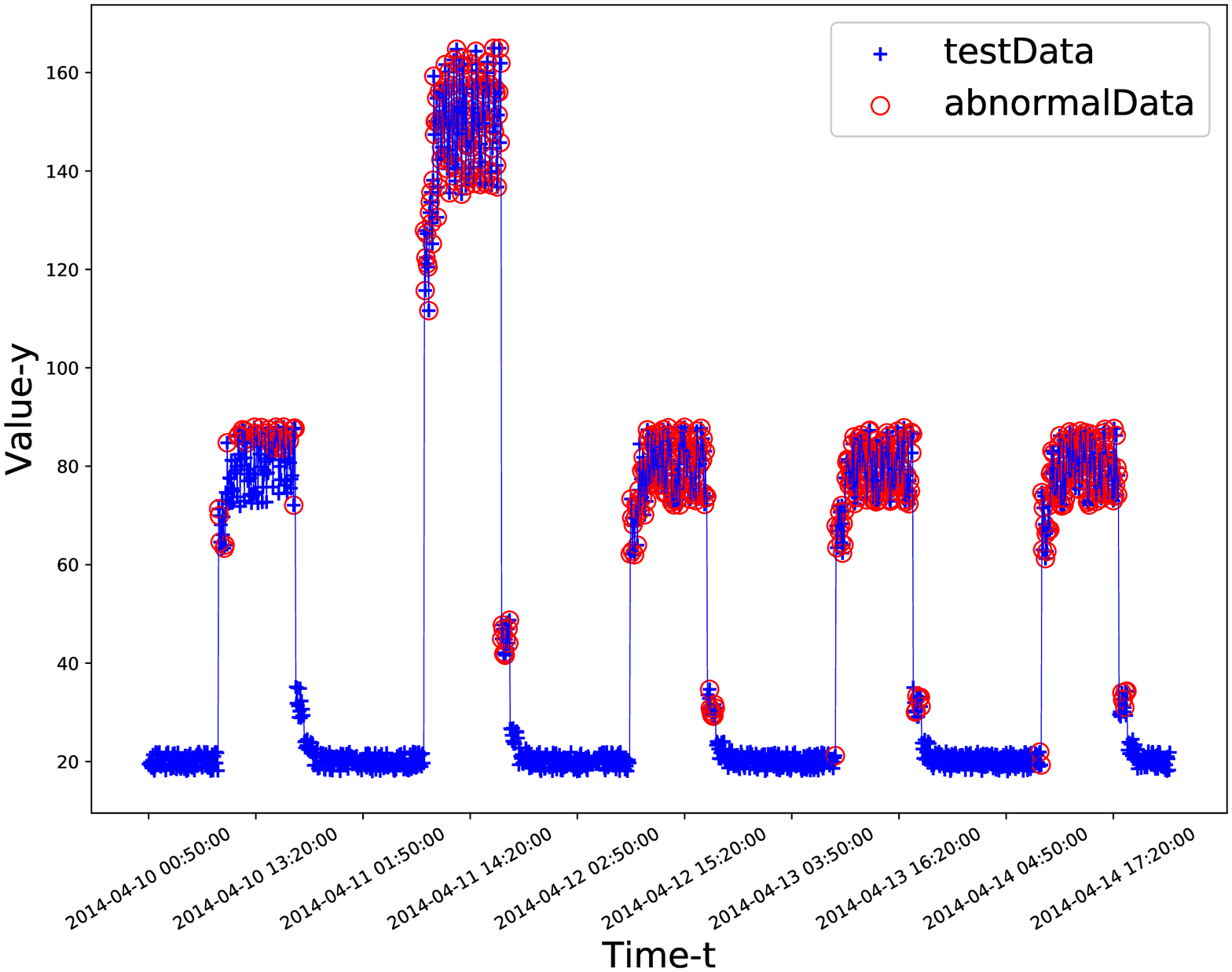}}
\subfigure[GPR-ADAM]
{\includegraphics[width=2.7in]{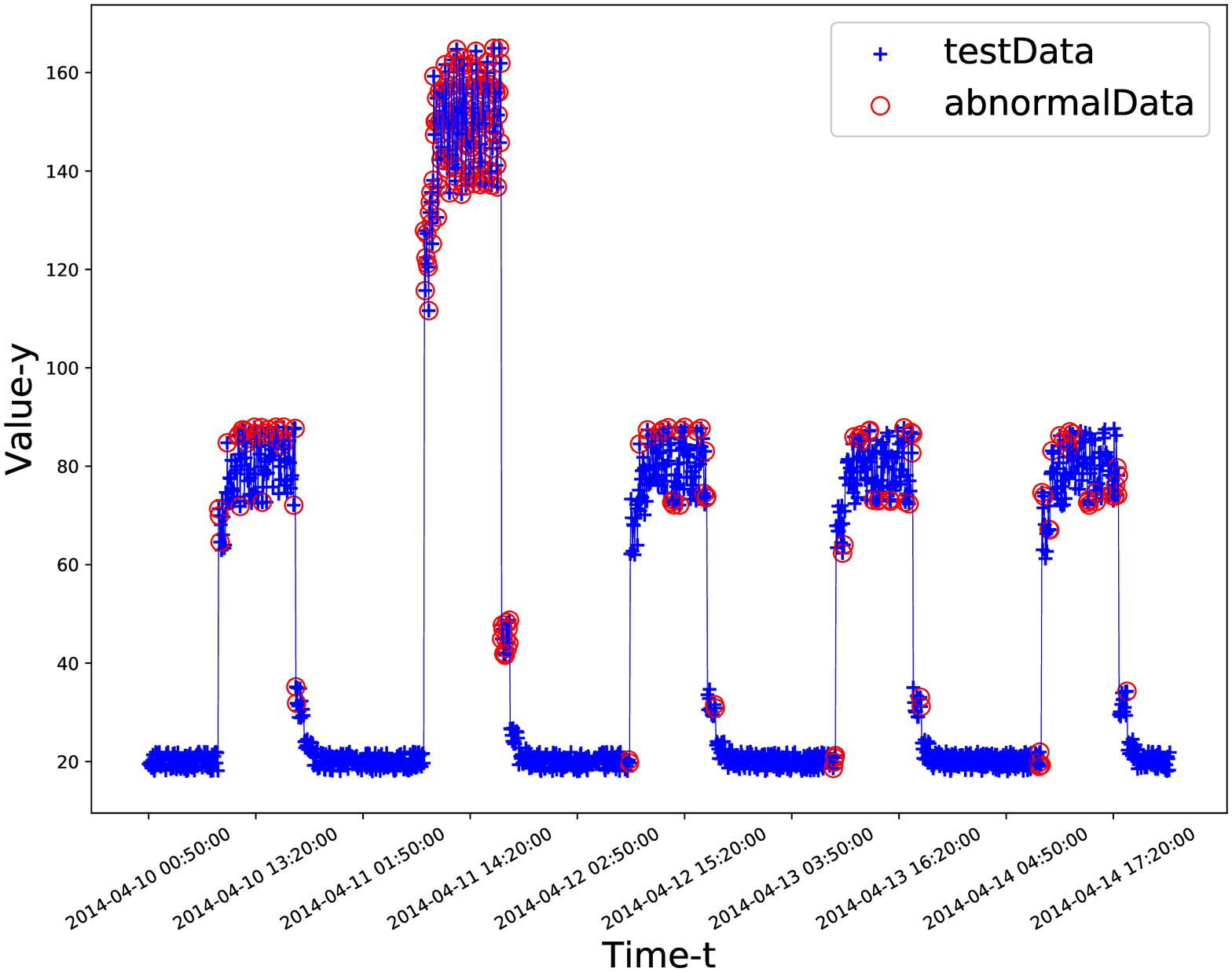}}
\subfigure[GPR-IADAM]
{\includegraphics[width=2.7in]{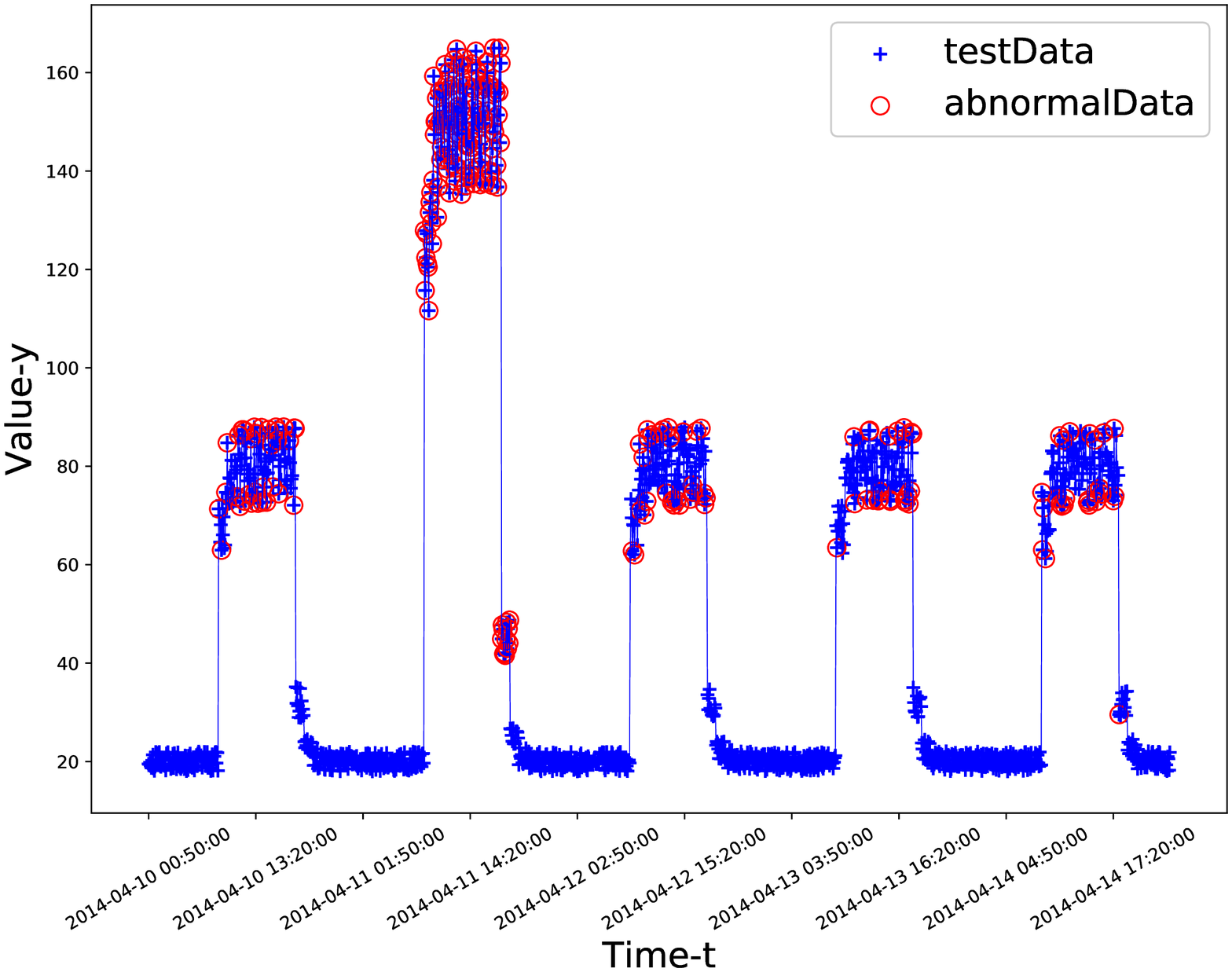}}
\subfigure[SGP-Q]
{\includegraphics[width=2.7in]{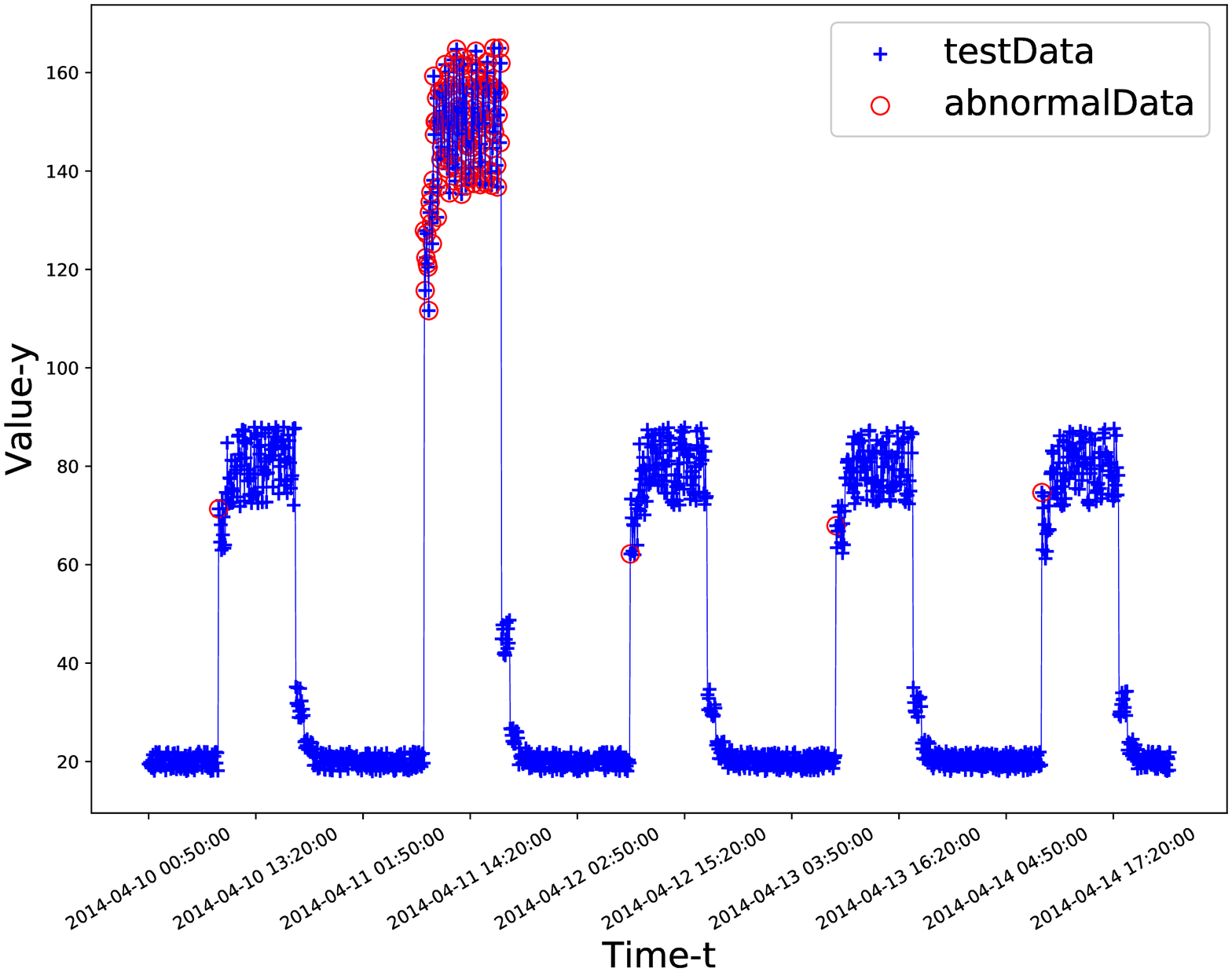}}
\subfigure[Label]
{\includegraphics[width=2.7in]{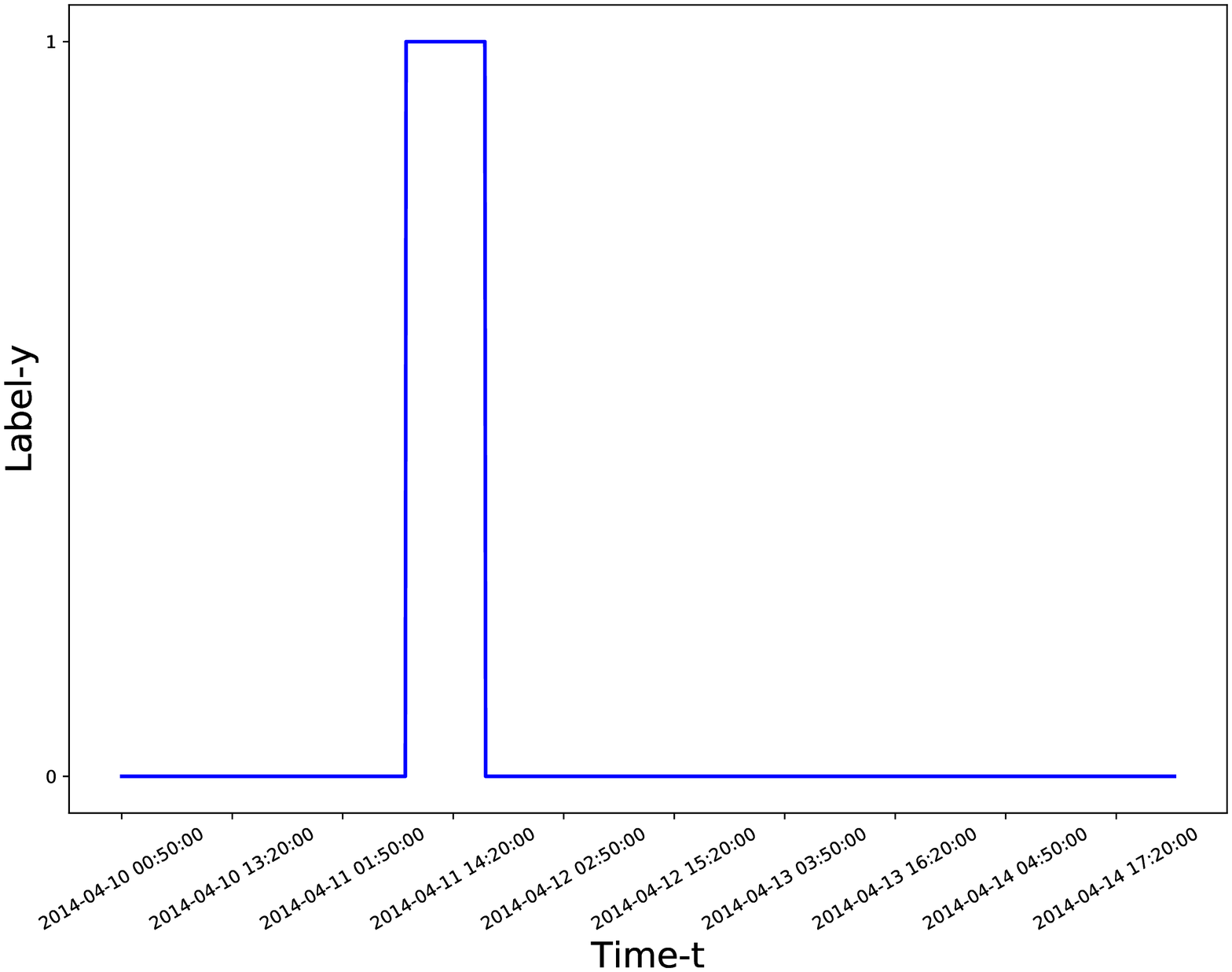}}
\caption{(a), (b), (c) and (d) show the results of anomaly detection of four methods on the `art\_daily\_jumpsup' dataset. (e) shows labels of test data, where 1 represents abnormal data and 0 represents normal data.}\label{fig:artdailyjumpsup}
\vspace{-0.1in}
\end{figure}

\begin{figure}[ht]
\centering
\subfigure[GPR-AD]
{\includegraphics[width=2.7in]{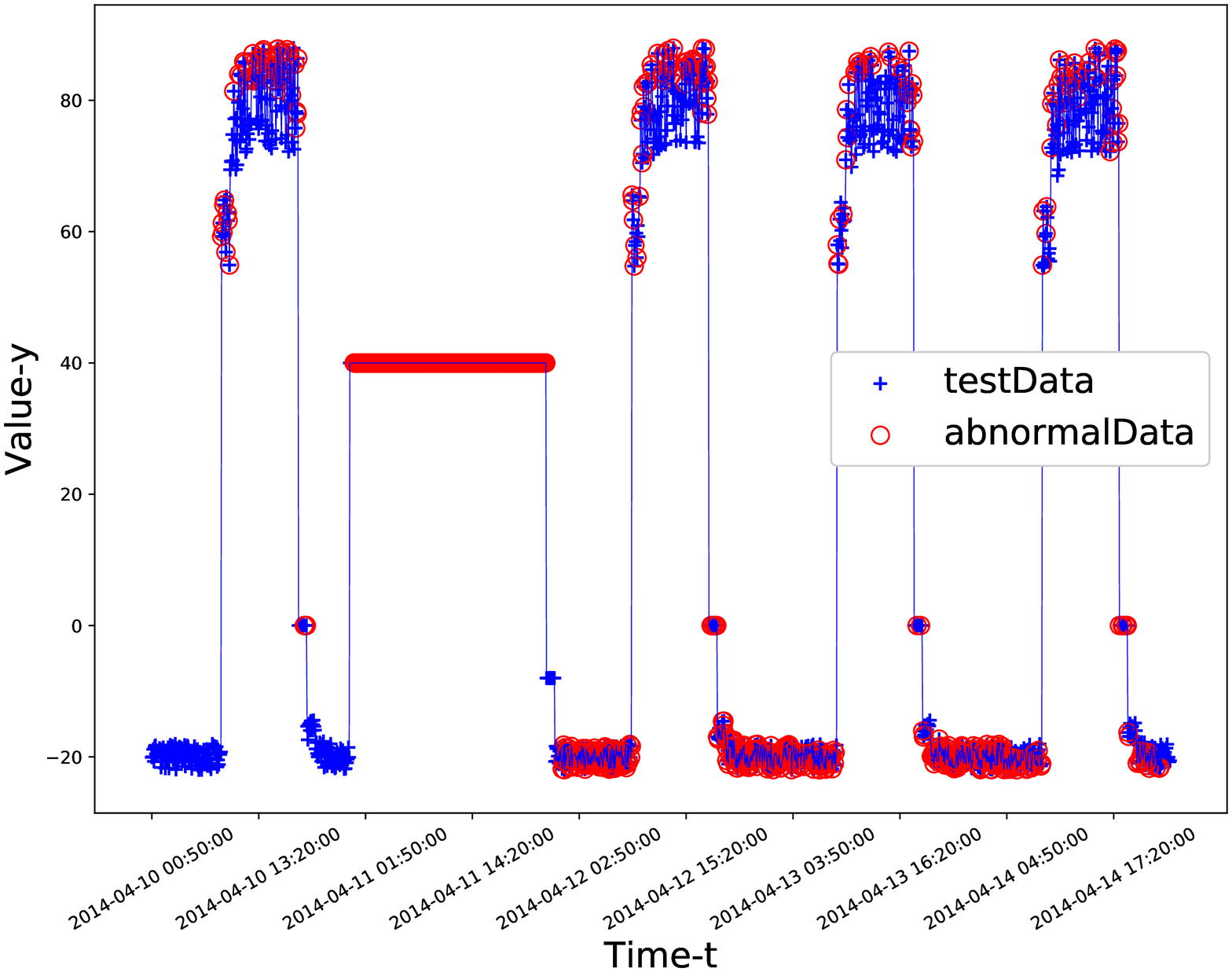}}
\subfigure[GPR-ADAM]
{\includegraphics[width=2.7in]{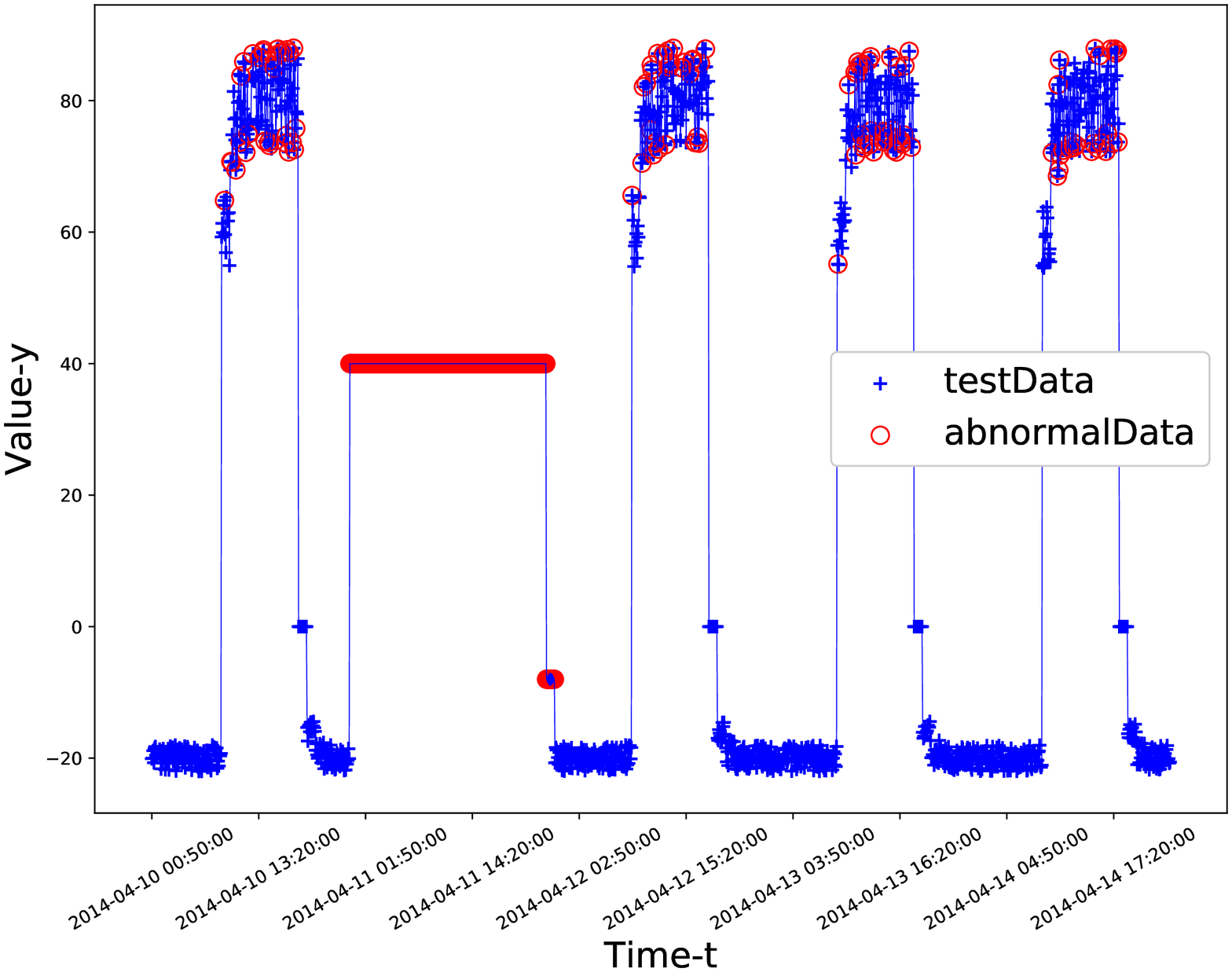}}
\subfigure[GPR-IADAM]
{\includegraphics[width=2.7in]{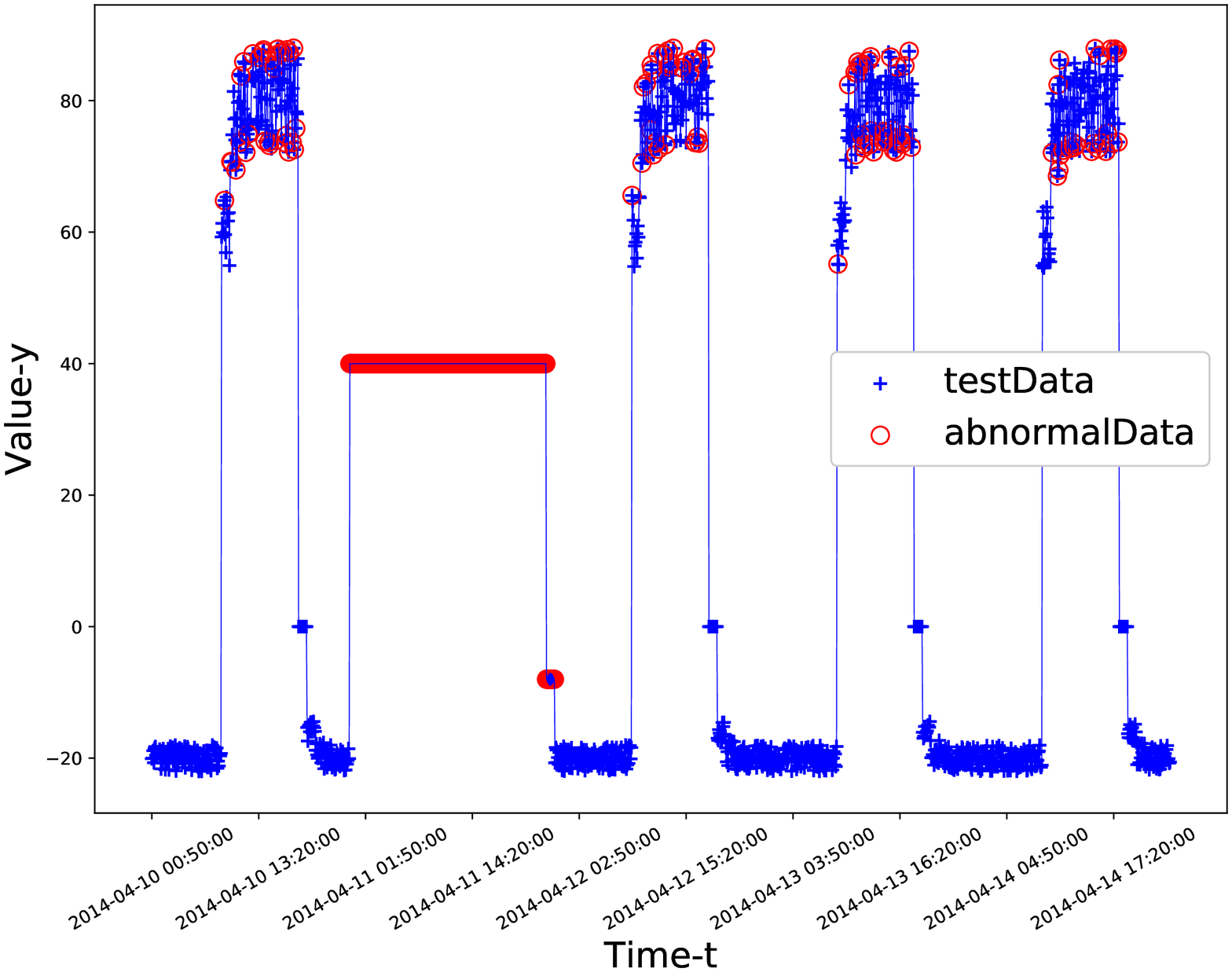}}
\subfigure[SGP-Q]
{\includegraphics[width=2.7in]{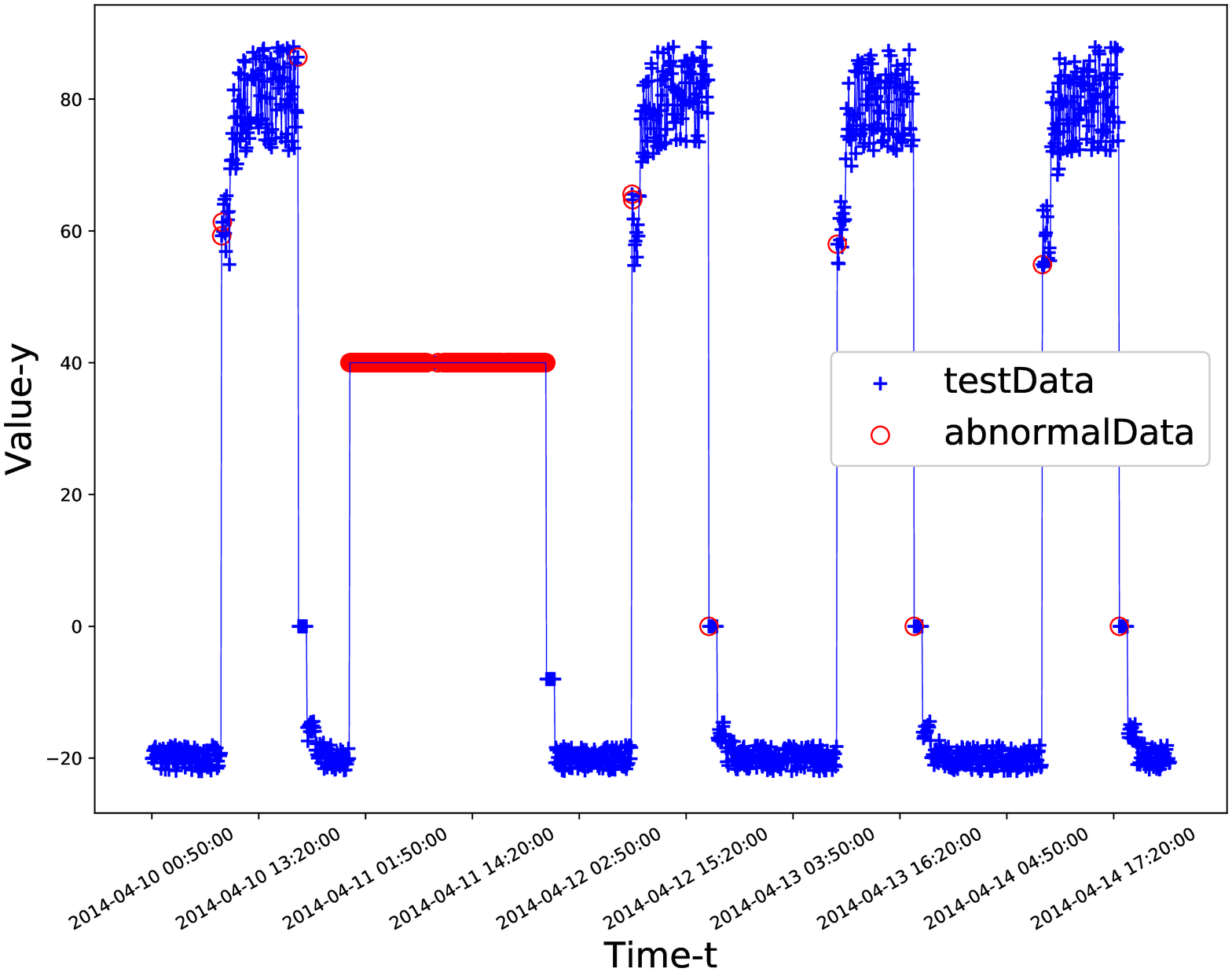}}
\subfigure[Label]
{\includegraphics[width=2.7in]{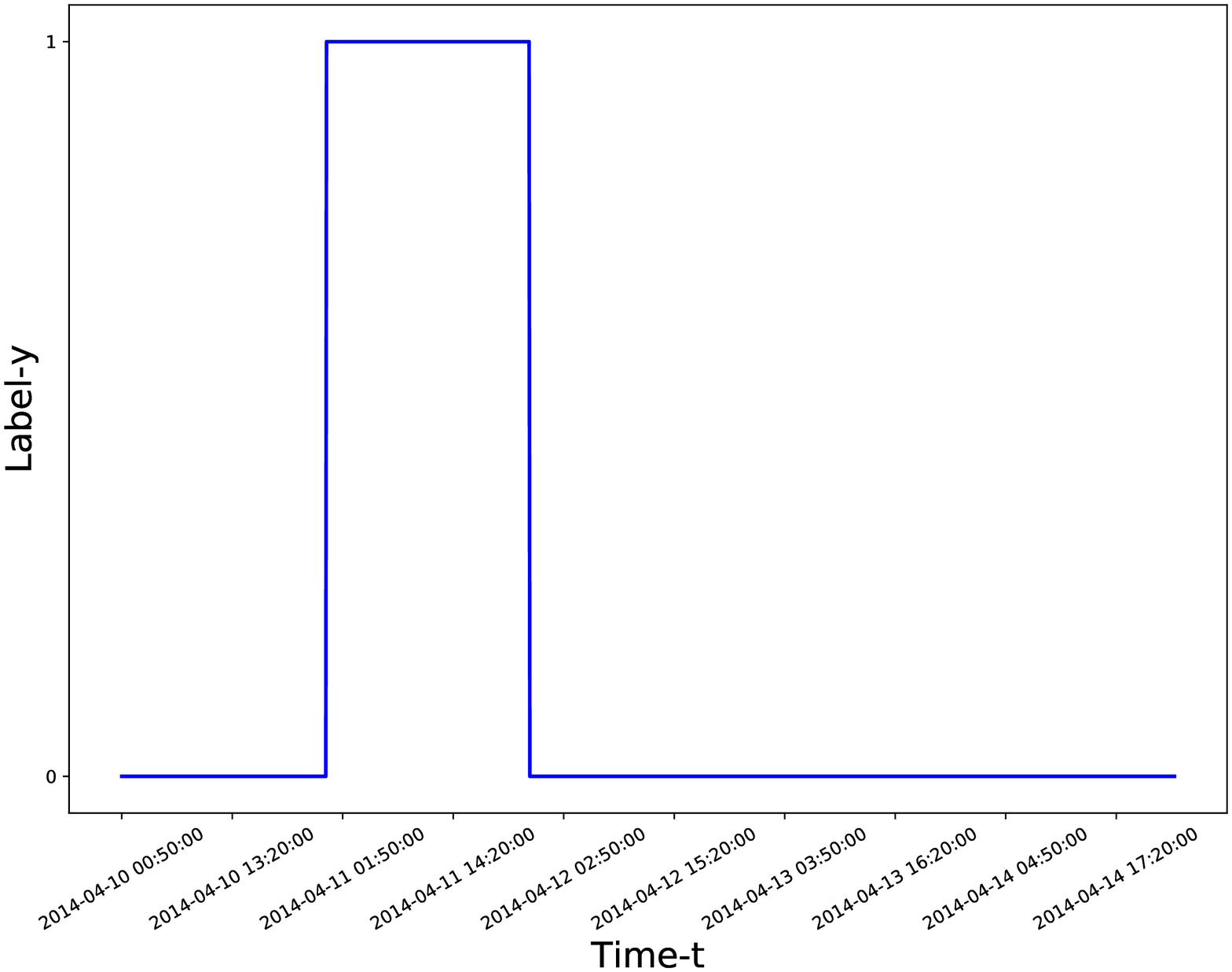}}
\caption{(a), (b), (c) and (d) show the results of anomaly detection of four methods on the `art\_daily\_flatmiddle' dataset. (e) shows labels of test data, where 1 represents abnormal data and 0 represents normal data.}\label{fig:artdailyflatmiddle}
\vspace{-0.1in}
\end{figure}

Figure \ref{fig:artdailyjumpsup} and Figure \ref{fig:artdailyflatmiddle} show the results of anomaly detection using four methods on the artificial datasets. As shown in Figure \ref{fig:artdailyjumpsup} (a) and Figure \ref{fig:artdailyflatmiddle} (a), the GPR-AD has the worst performance. The reason is that the GPR-AD adds lots of abnormal data to the training data, which leads to inaccurate training models and marks subsequent normal data as abnormal data. In Figure \ref{fig:artdailyjumpsup} (b), (c) and Figure \ref{fig:artdailyflatmiddle} (b), (c), the GPR-ADAM and GPR-IADAM have similar performance when abnormal data deviate from normal data obviously. Figure \ref{fig:artdailyjumpsup} (d) and Figure \ref{fig:artdailyflatmiddle} (d) show that our method SGP-Q has the best performance. Almost all abnormal data are detected, and only a few normal data are wrongly marked as abnormal data by the SGP-Q.

\begin{figure}[ht]
\centering
\subfigure[GPR-AD]
{\includegraphics[width=2.7in]{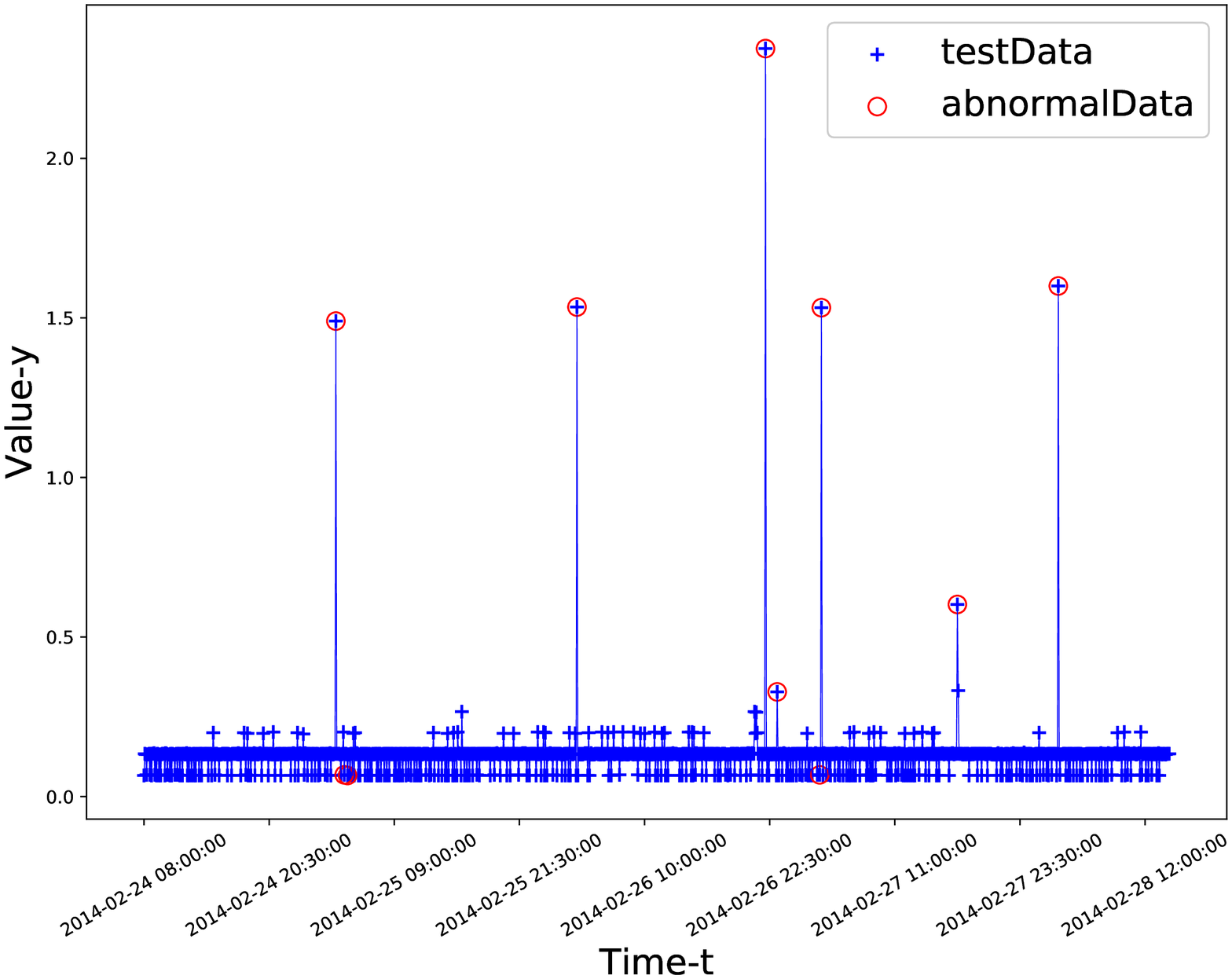}}
\subfigure[GPR-ADAM]
{\includegraphics[width=2.7in]{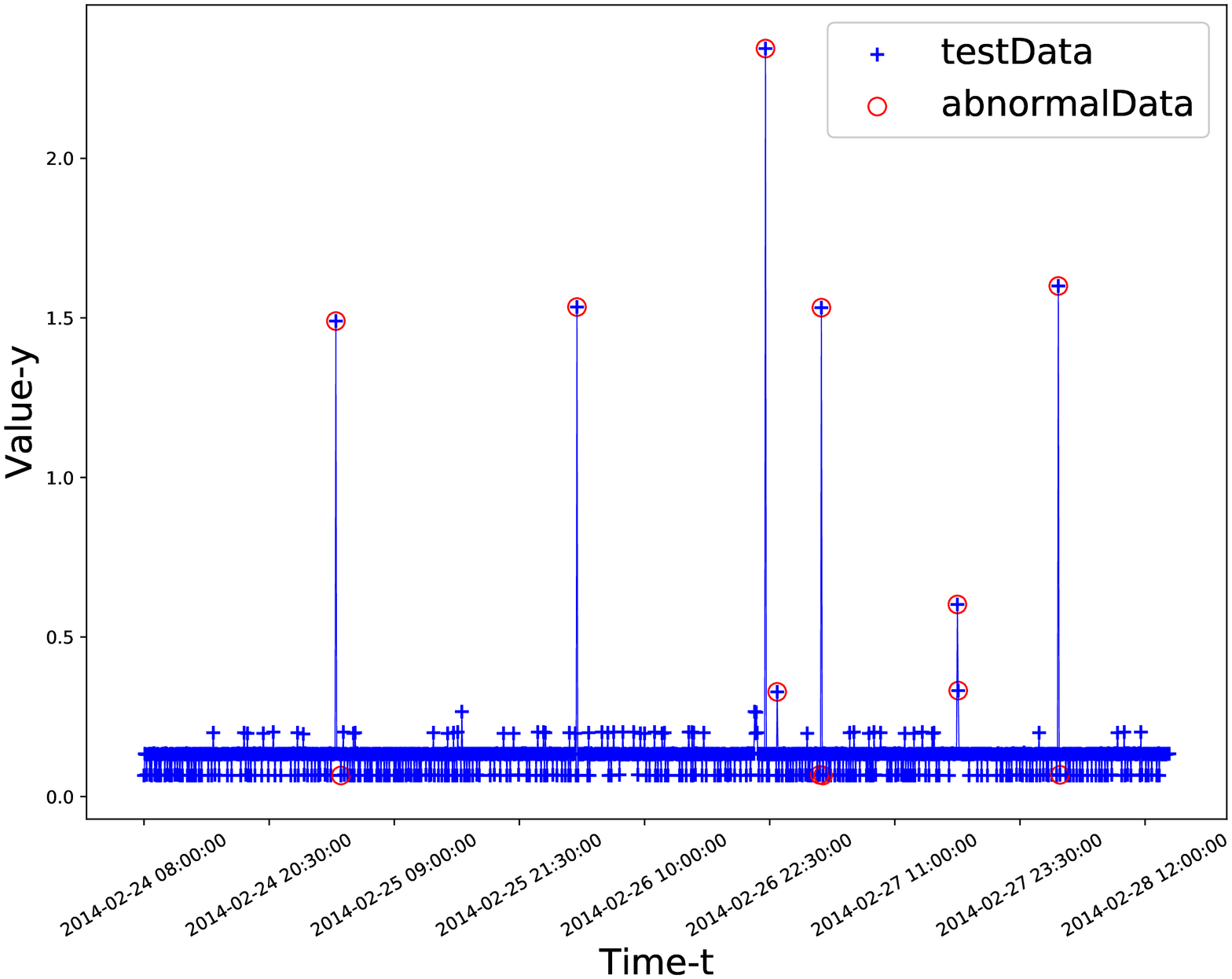}}
\subfigure[GPR-IADAM]
{\includegraphics[width=2.7in]{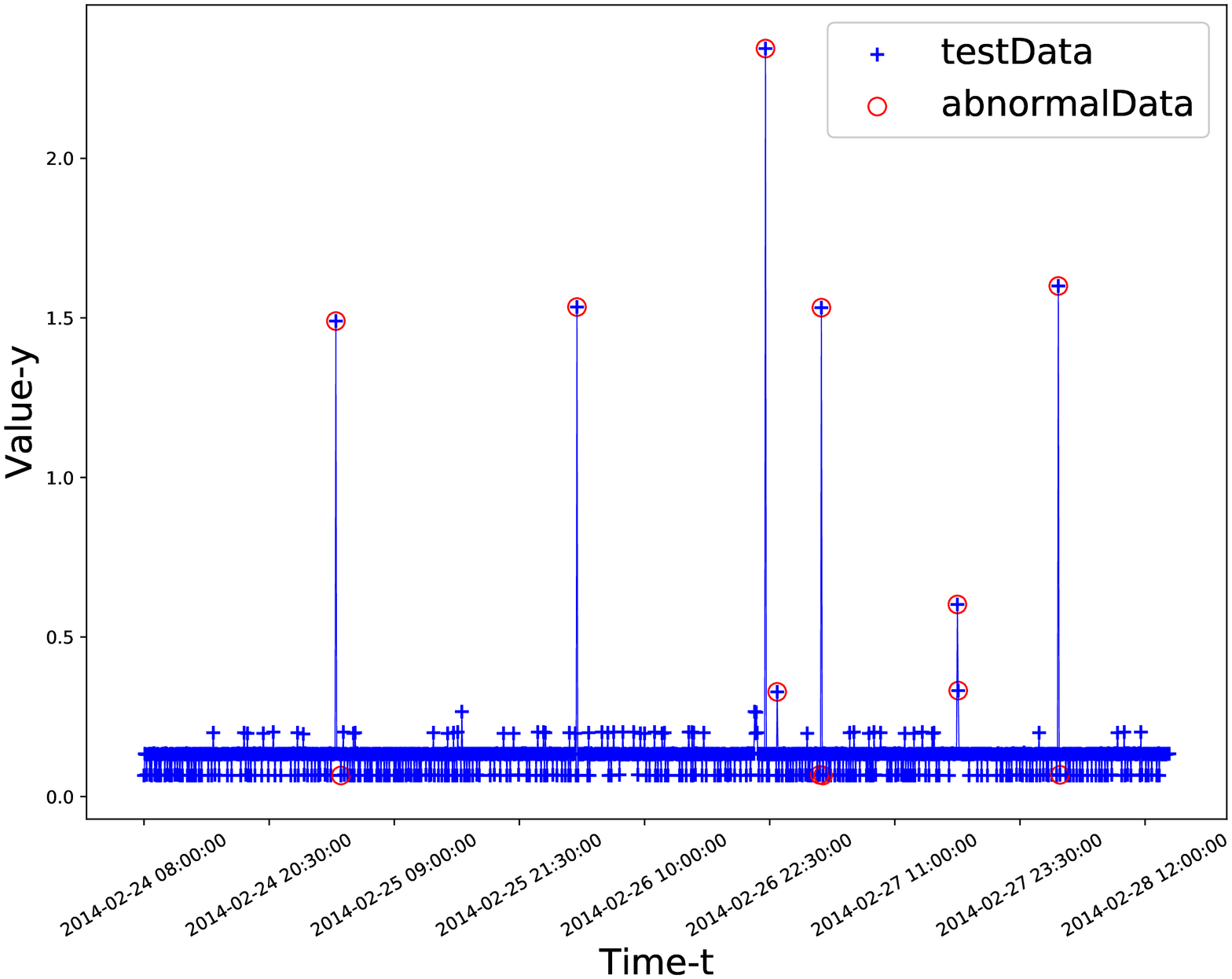}}
\subfigure[SGP-Q]
{\includegraphics[width=2.7in]{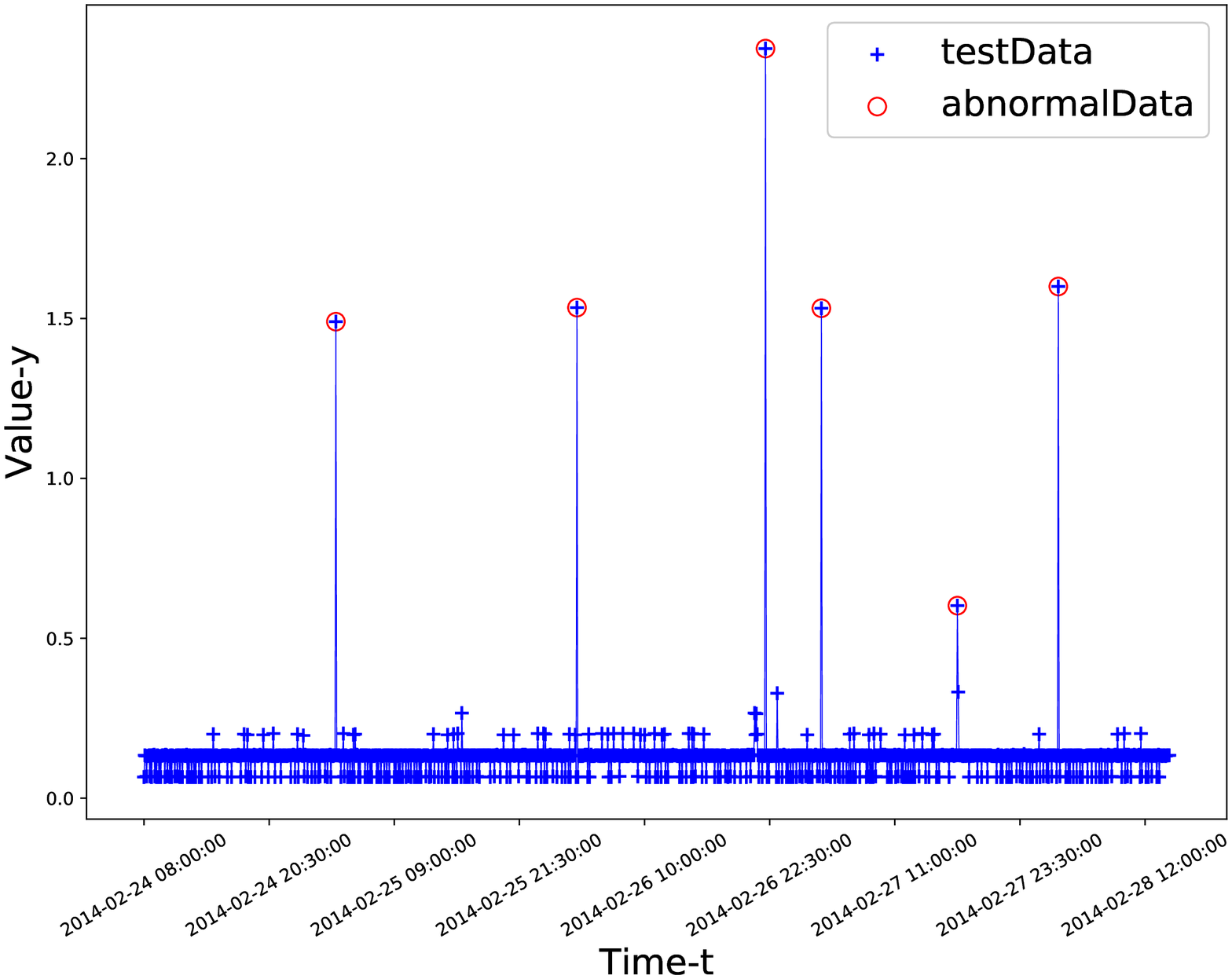}}
\subfigure[Label]
{\includegraphics[width=2.7in]{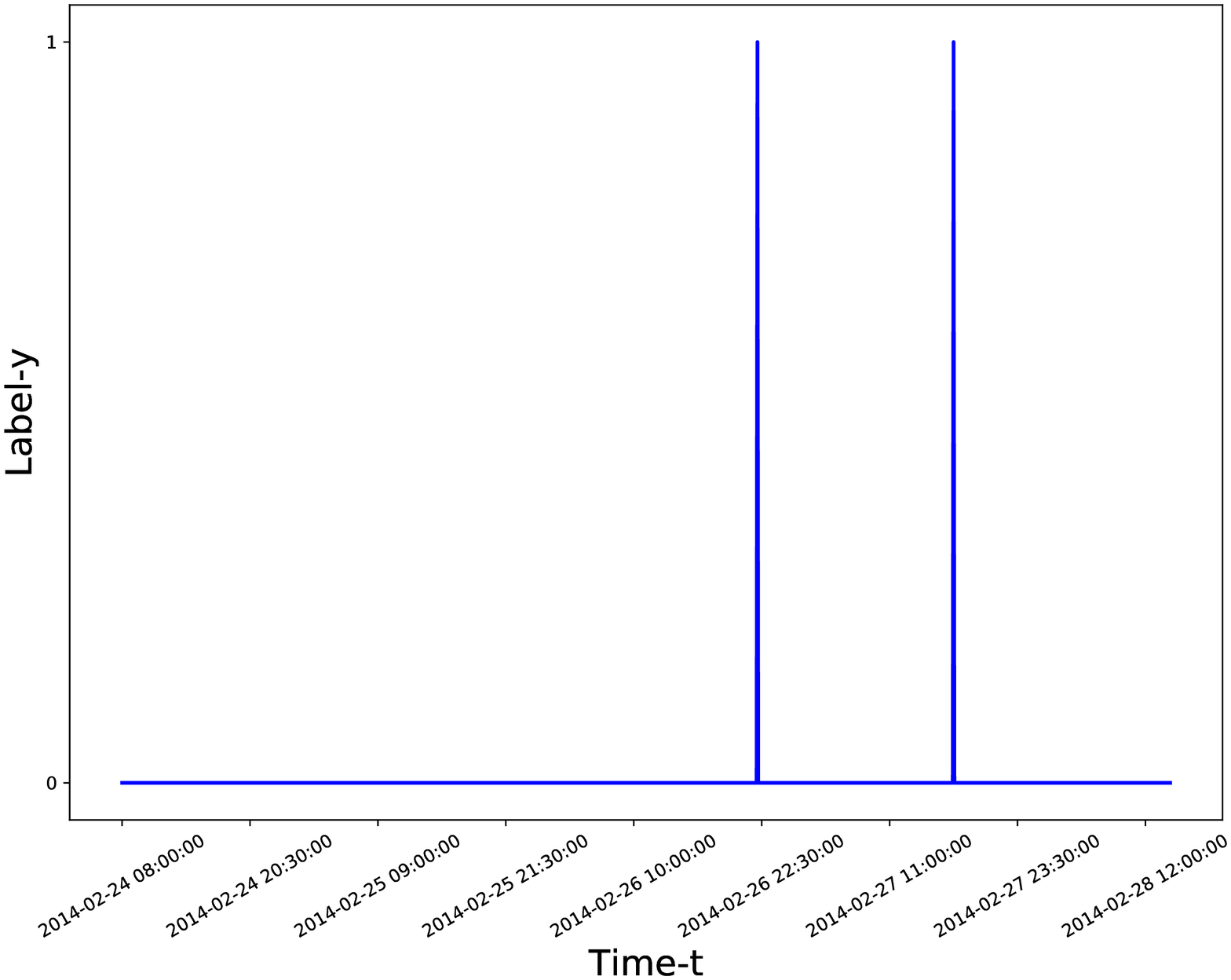}}
\caption{(a), (b), (c) and (d) show the results of anomaly detection of four methods on the `ec2\_cpu\_utilization\_24ae8d' dataset. (e) shows labels of test data, where 1 represents abnormal data and 0 represents normal data.}\label{fig:ec2cpuutilization24ae8d}
\vspace{-0.1in}
\end{figure}

\begin{figure}[ht]
\centering
\subfigure[GPR-AD]
{\includegraphics[width=2.7in]{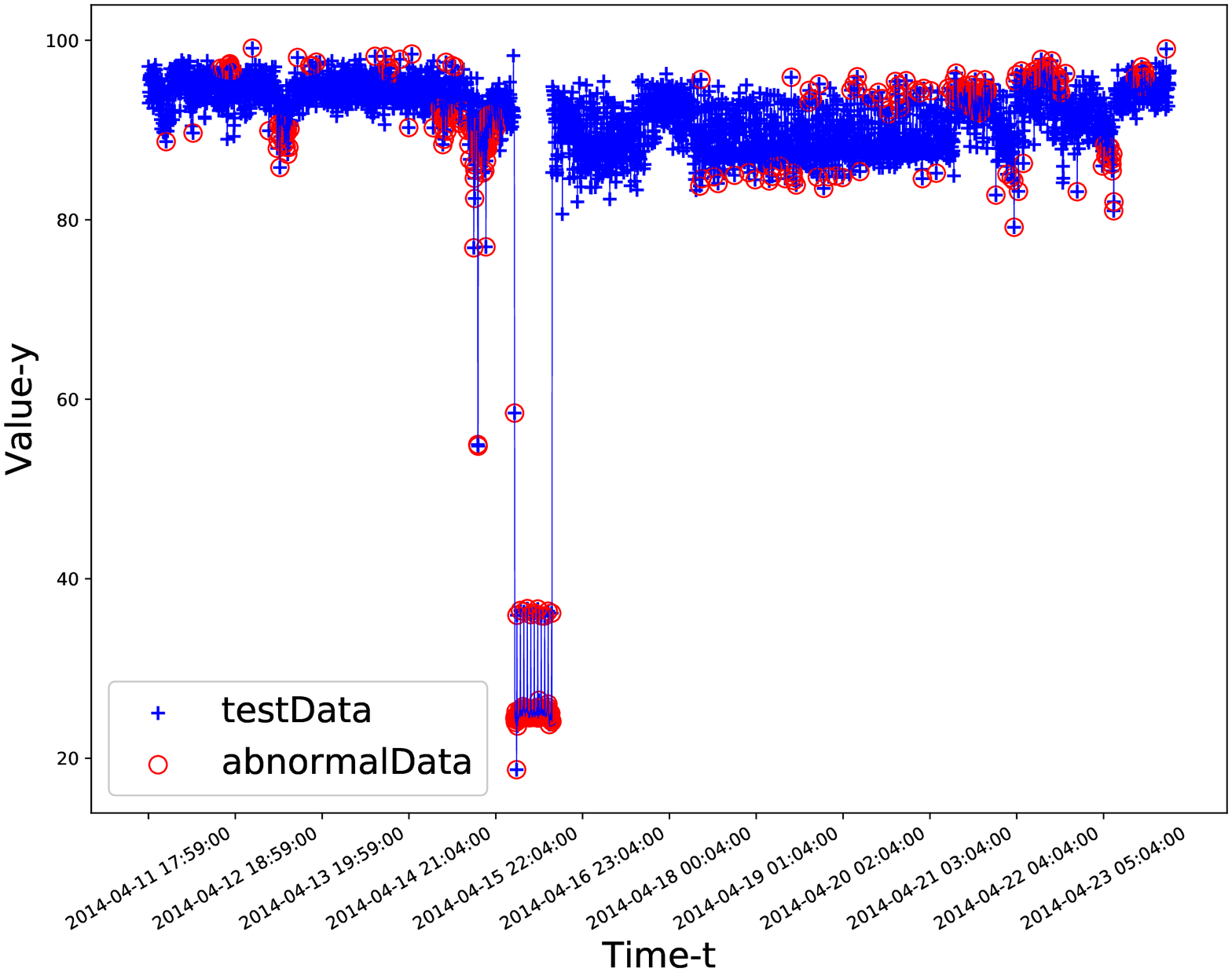}}
\subfigure[GPR-ADAM]
{\includegraphics[width=2.7in]{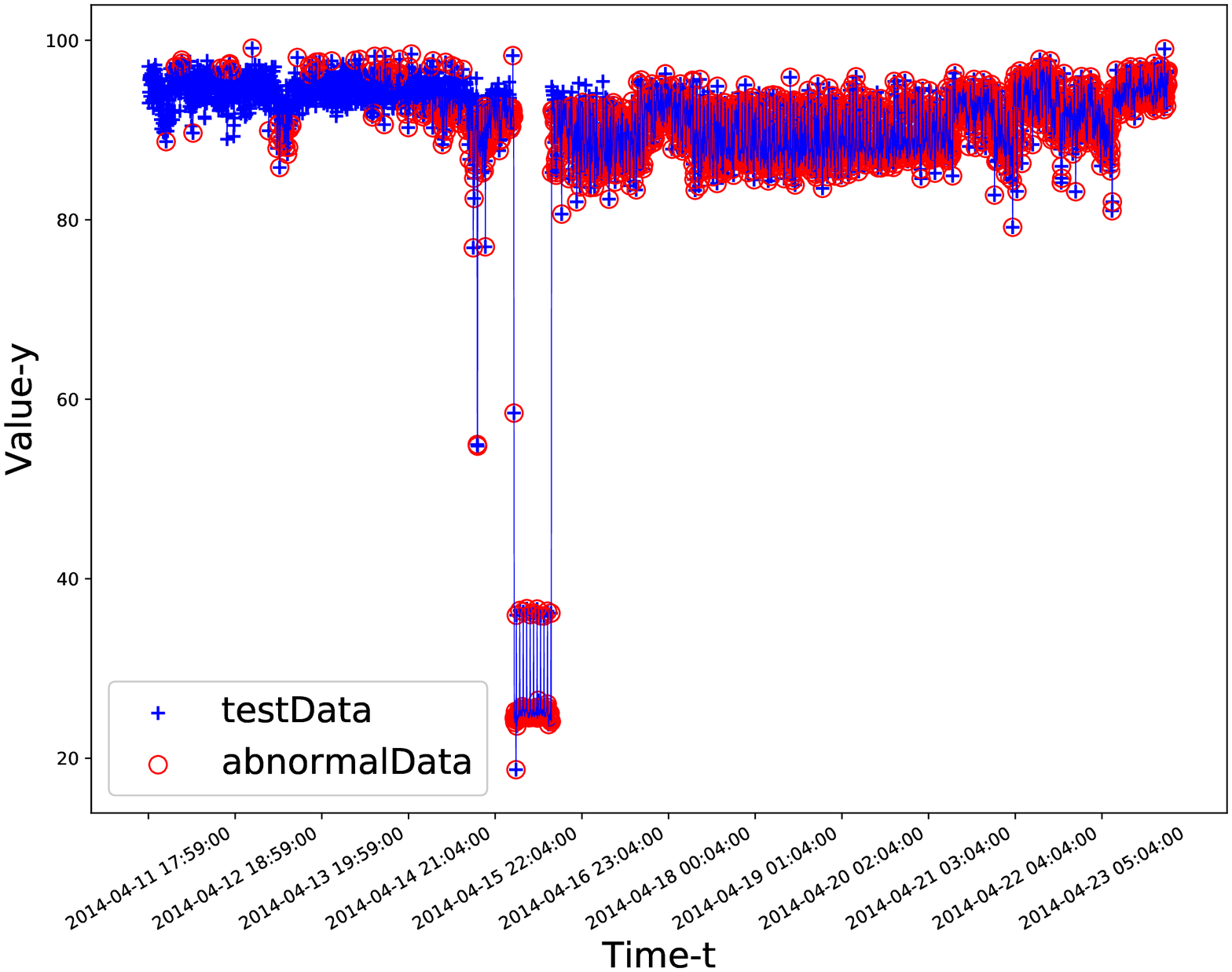}}
\subfigure[GPR-IADAM]
{\includegraphics[width=2.7in]{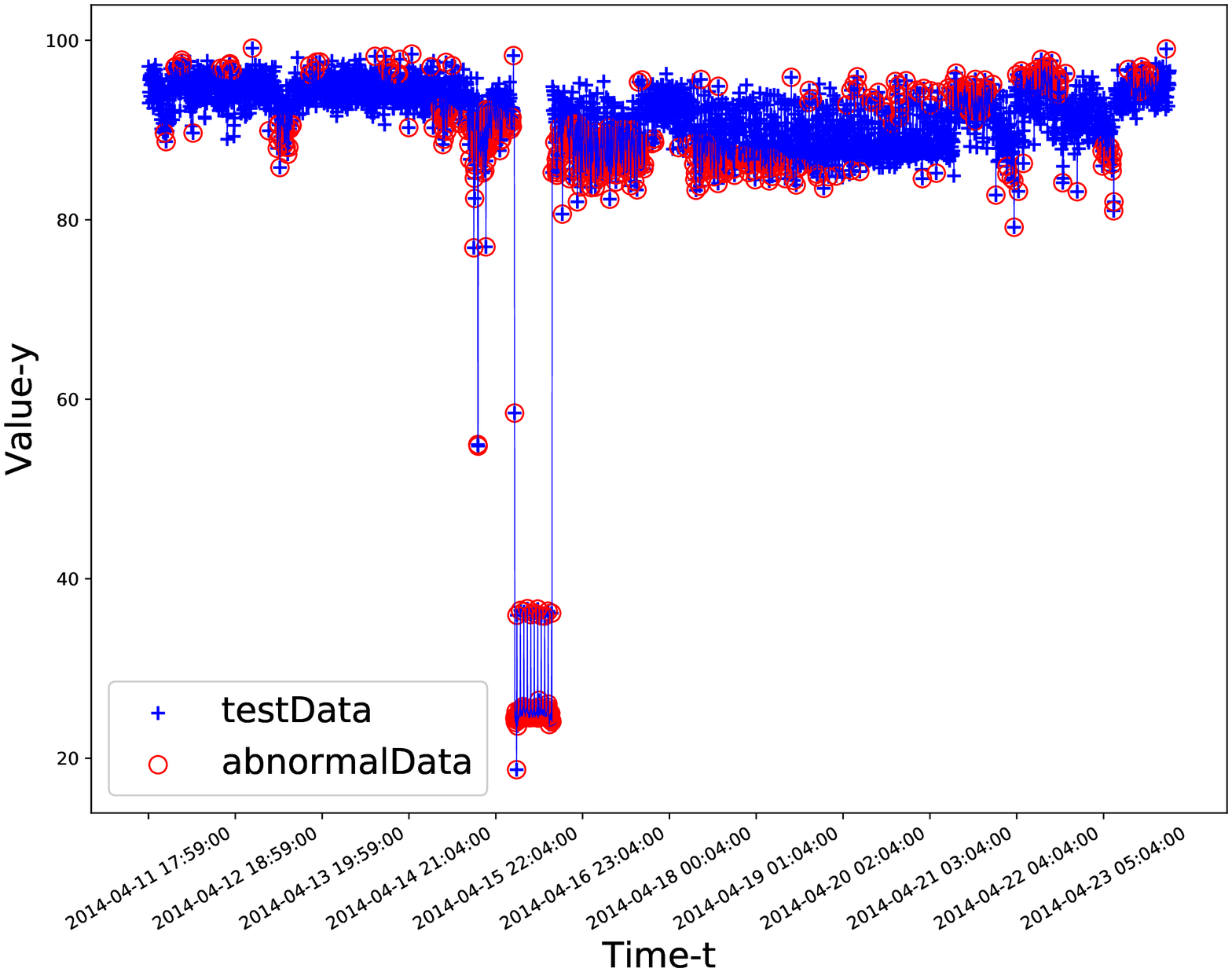}}
\subfigure[SGP-Q]
{\includegraphics[width=2.7in]{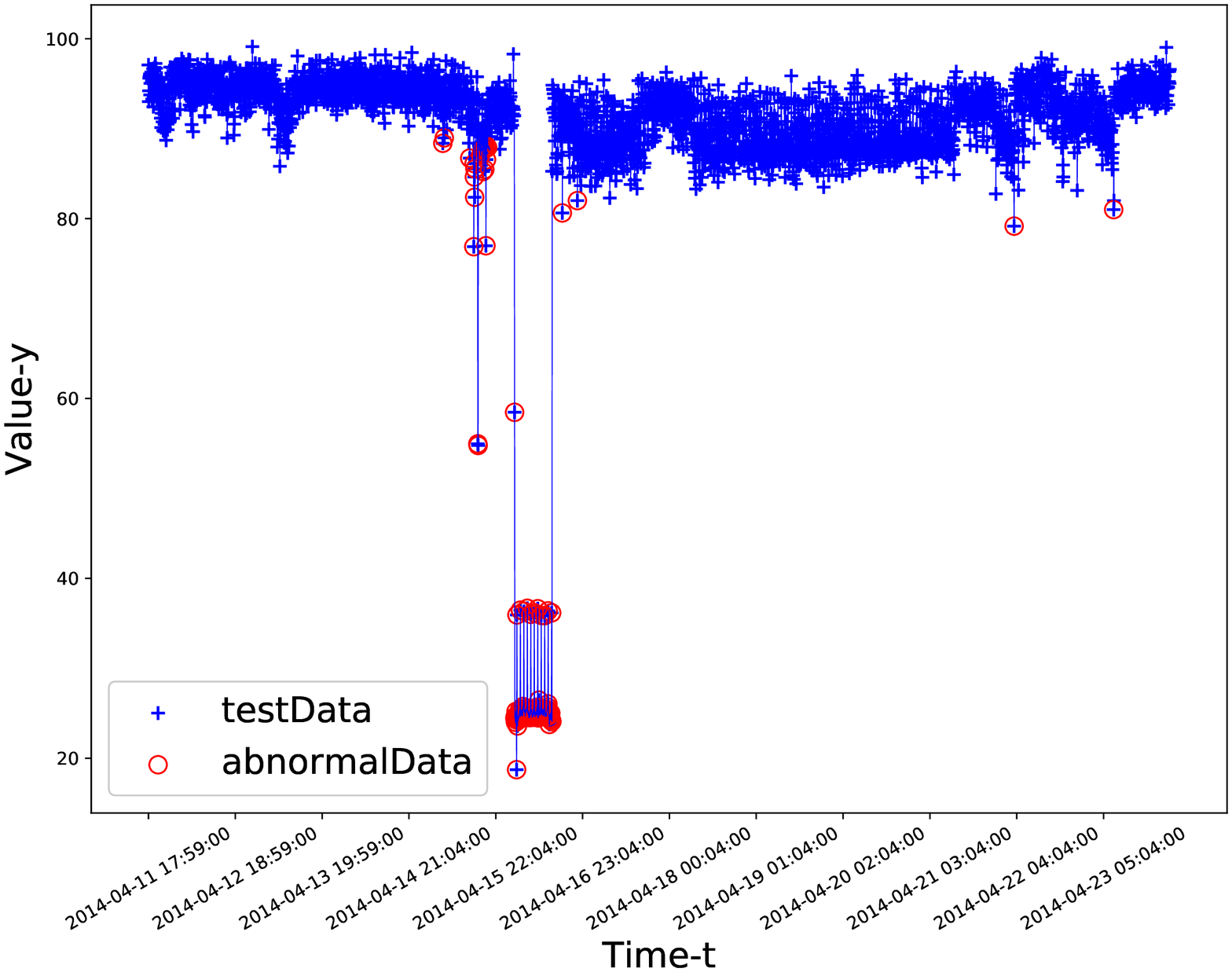}}
\subfigure[Label]
{\includegraphics[width=2.7in]{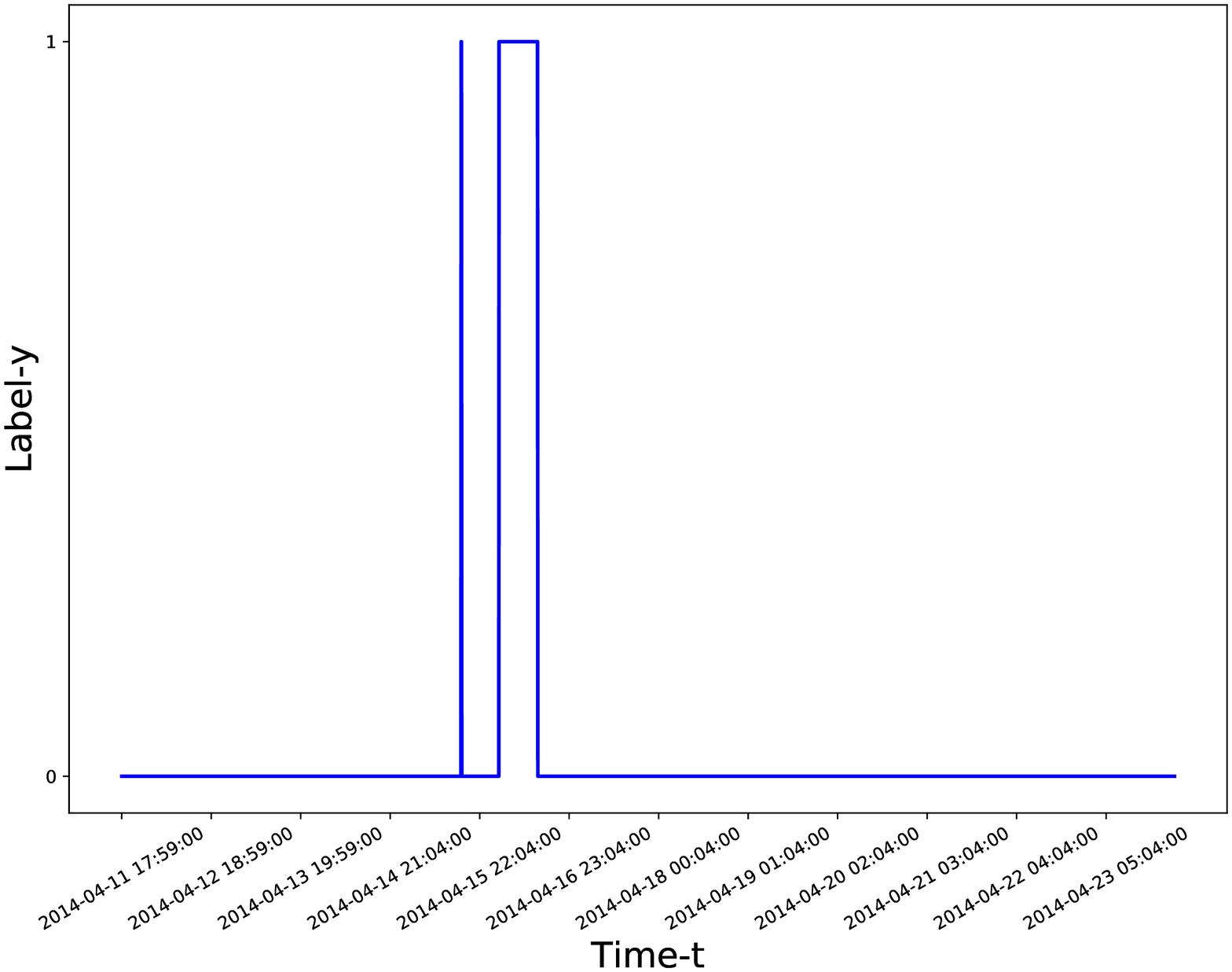}}
\caption{(a), (b), (c) and (d) show the results of anomaly detection of four methods on the `ec2\_cpu\_utilization\_825cc2' dataset. (e) shows labels of test data, where 1 represents abnormal data and 0 represents normal data.}\label{fig:ec2cpuutilization825cc2}
\vspace{-0.1in}
\end{figure}

\begin{figure}[ht]
\centering
\subfigure[GPR-AD]
{\includegraphics[width=2.7in]{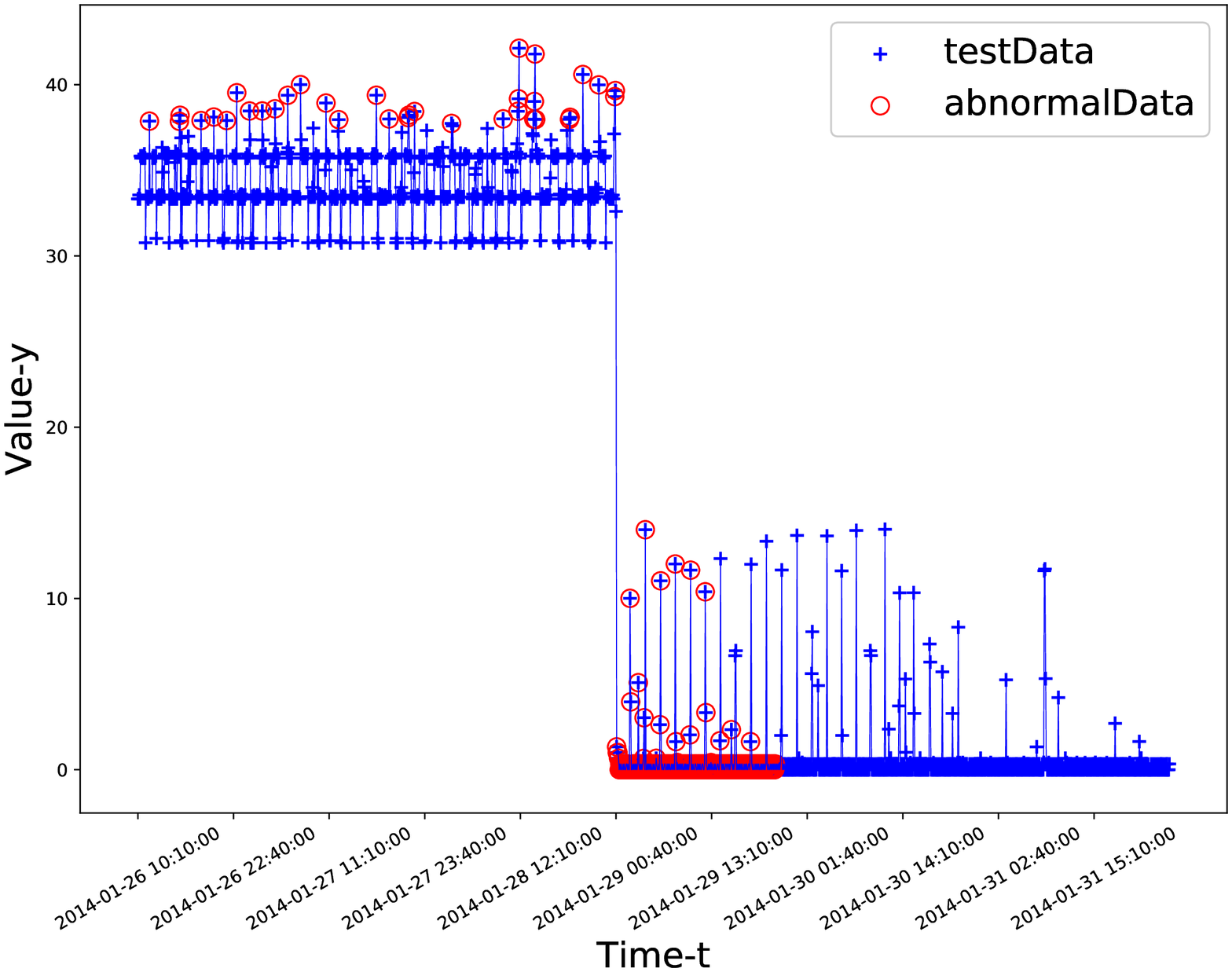}}
\subfigure[GPR-ADAM]
{\includegraphics[width=2.7in]{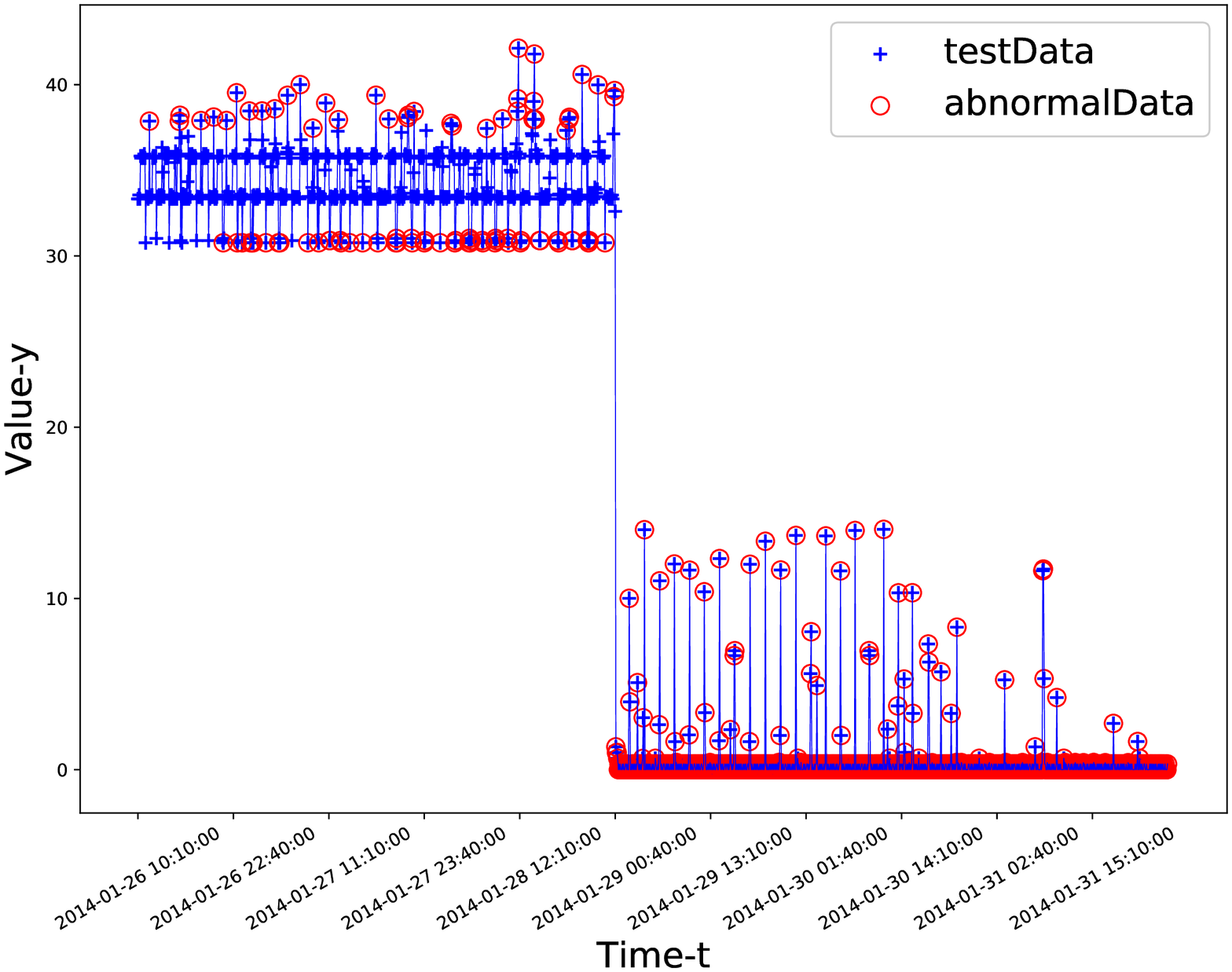}}
\subfigure[GPR-IADAM]
{\includegraphics[width=2.7in]{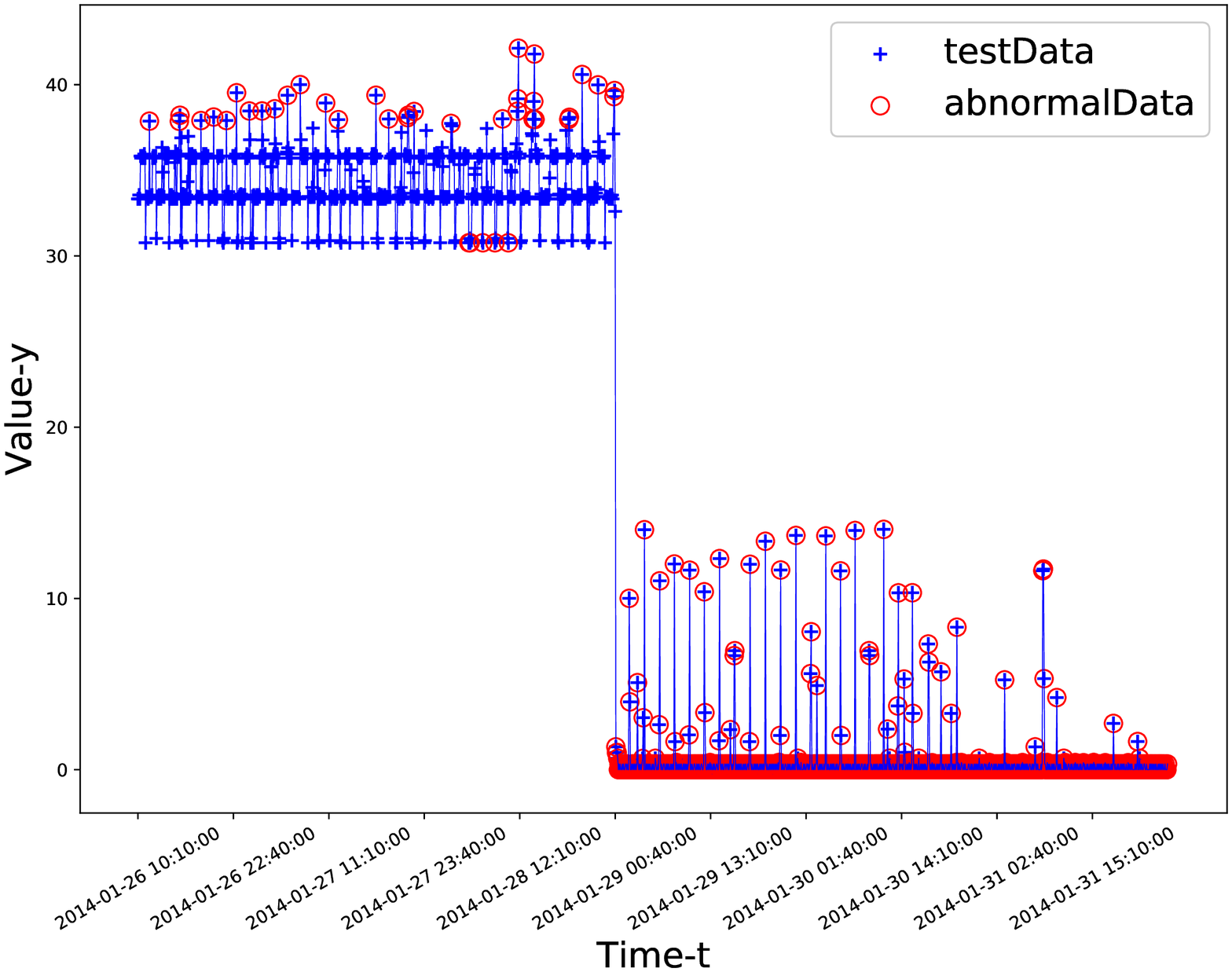}}
\subfigure[SGP-Q]
{\includegraphics[width=2.7in]{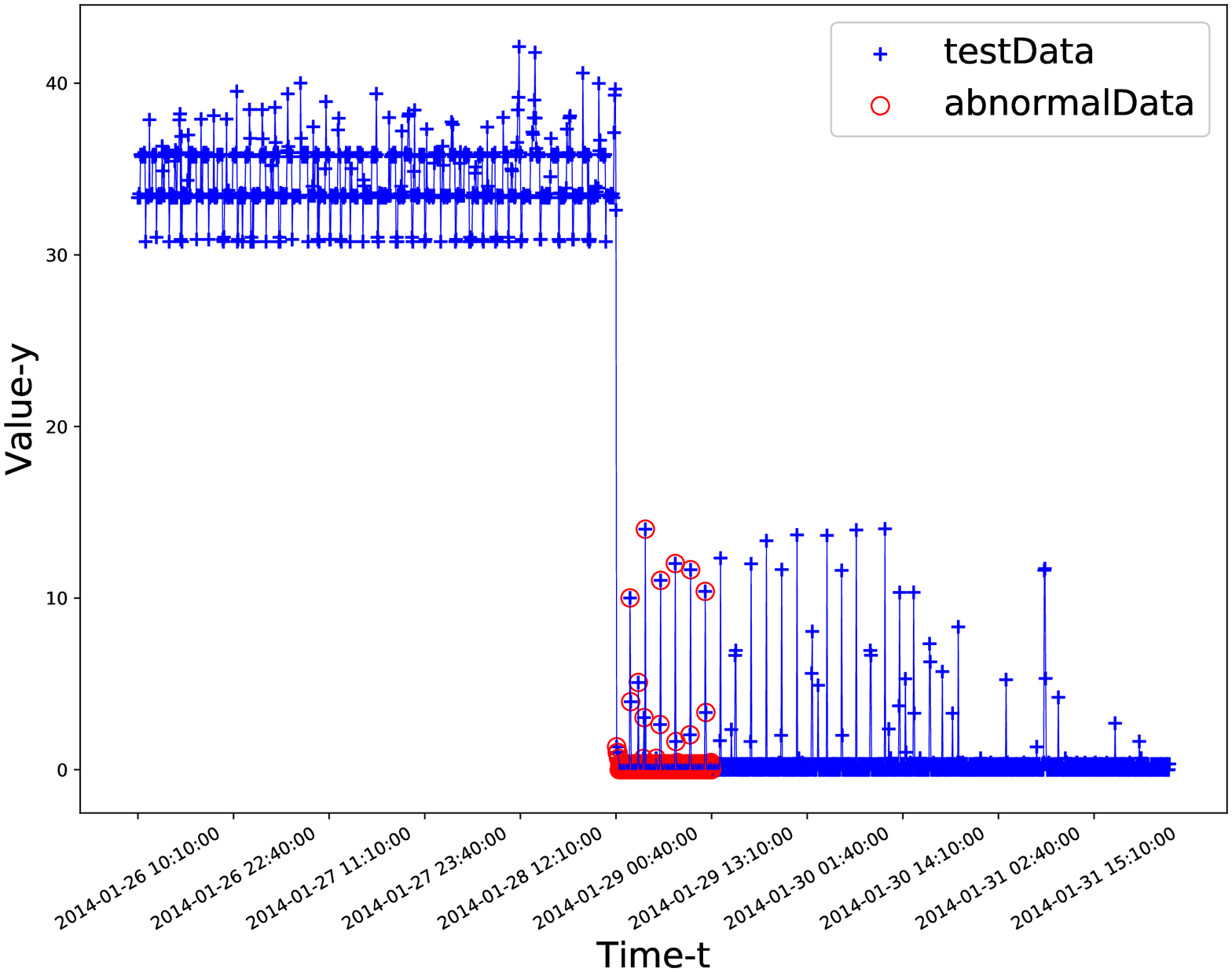}}
\subfigure[Label]
{\includegraphics[width=2.7in]{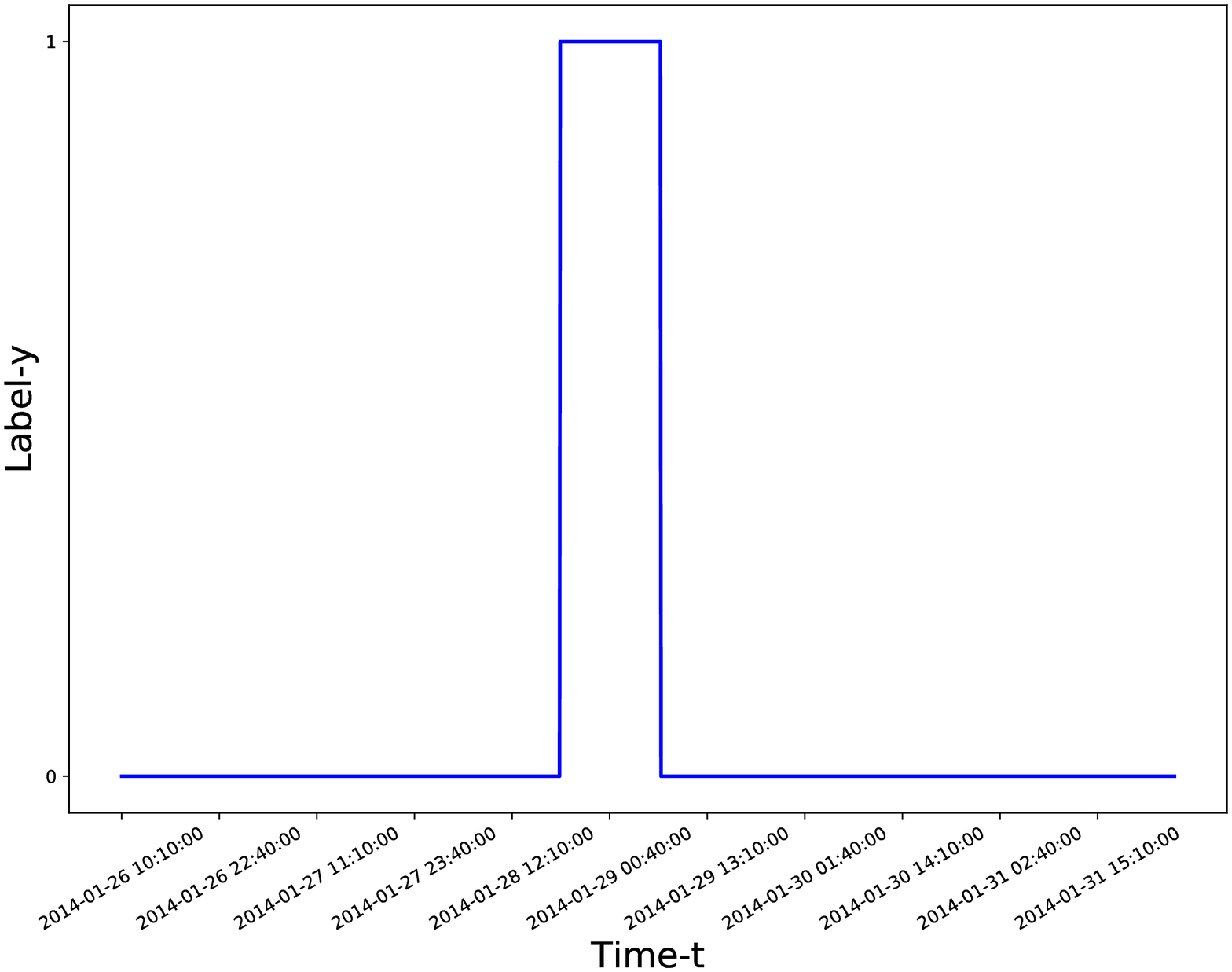}}
\caption{(a), (b), (c) and (d) show the results of anomaly detection of four methods on the `grok\_asg\_anomaly' dataset. (e) shows labels of test data, where 1 represents abnormal data and 0 represents normal data.}\label{fig:grokasganomaly}
\vspace{-0.1in}
\end{figure}

\begin{figure}[ht]
\centering
\subfigure[GPR-AD]
{\includegraphics[width=2.7in]{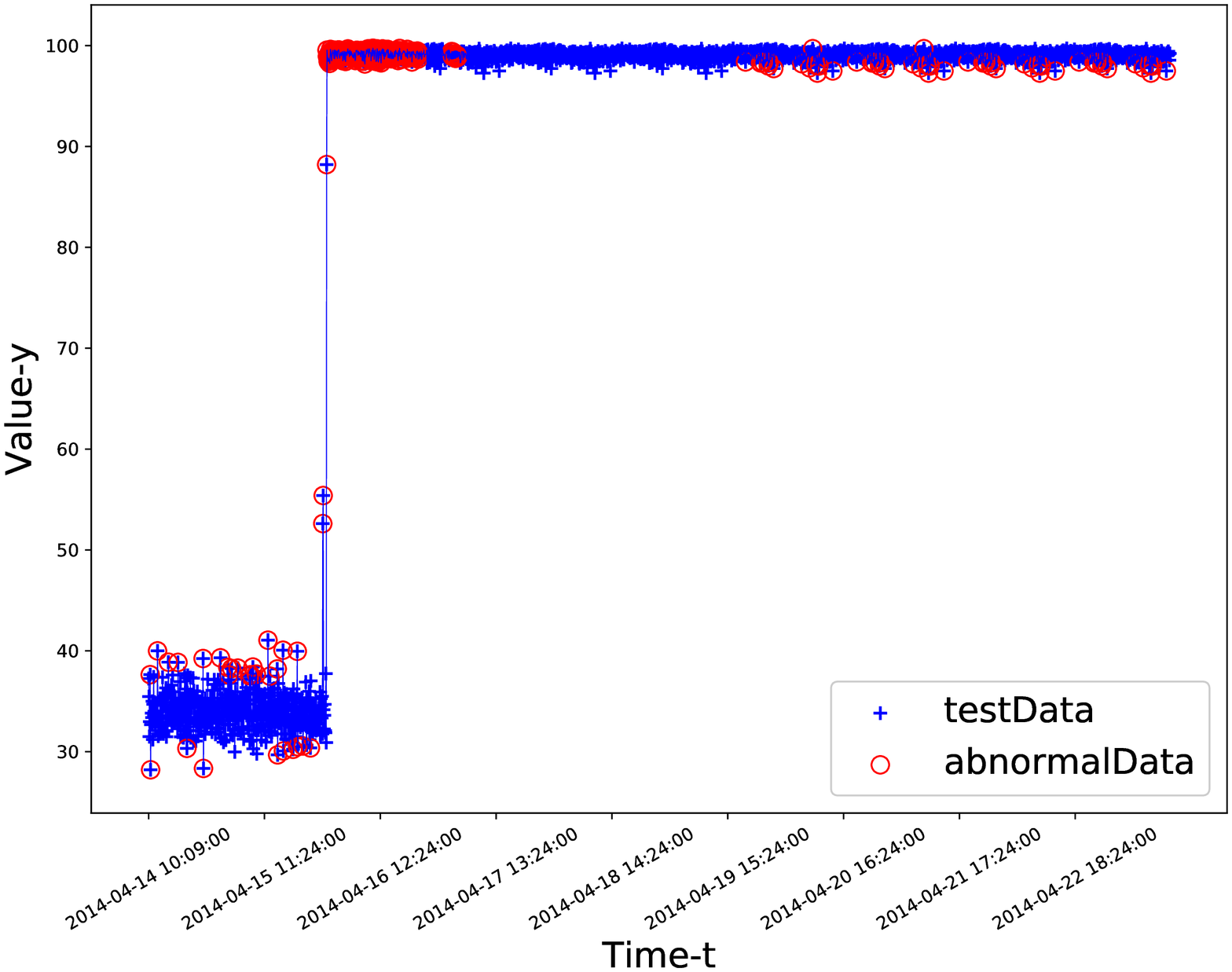}}
\subfigure[GPR-ADAM]
{\includegraphics[width=2.7in]{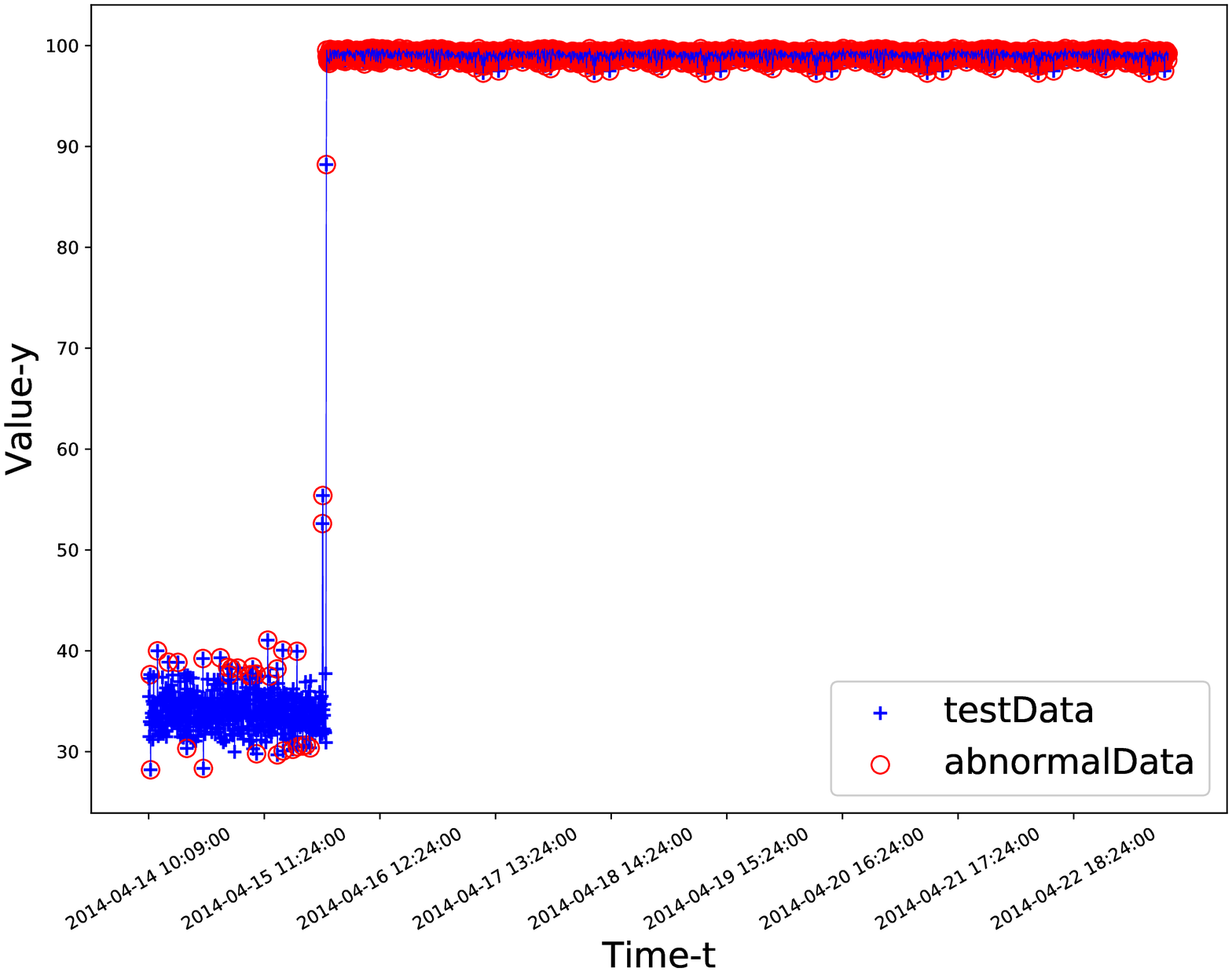}}
\subfigure[GPR-IADAM]
{\includegraphics[width=2.7in]{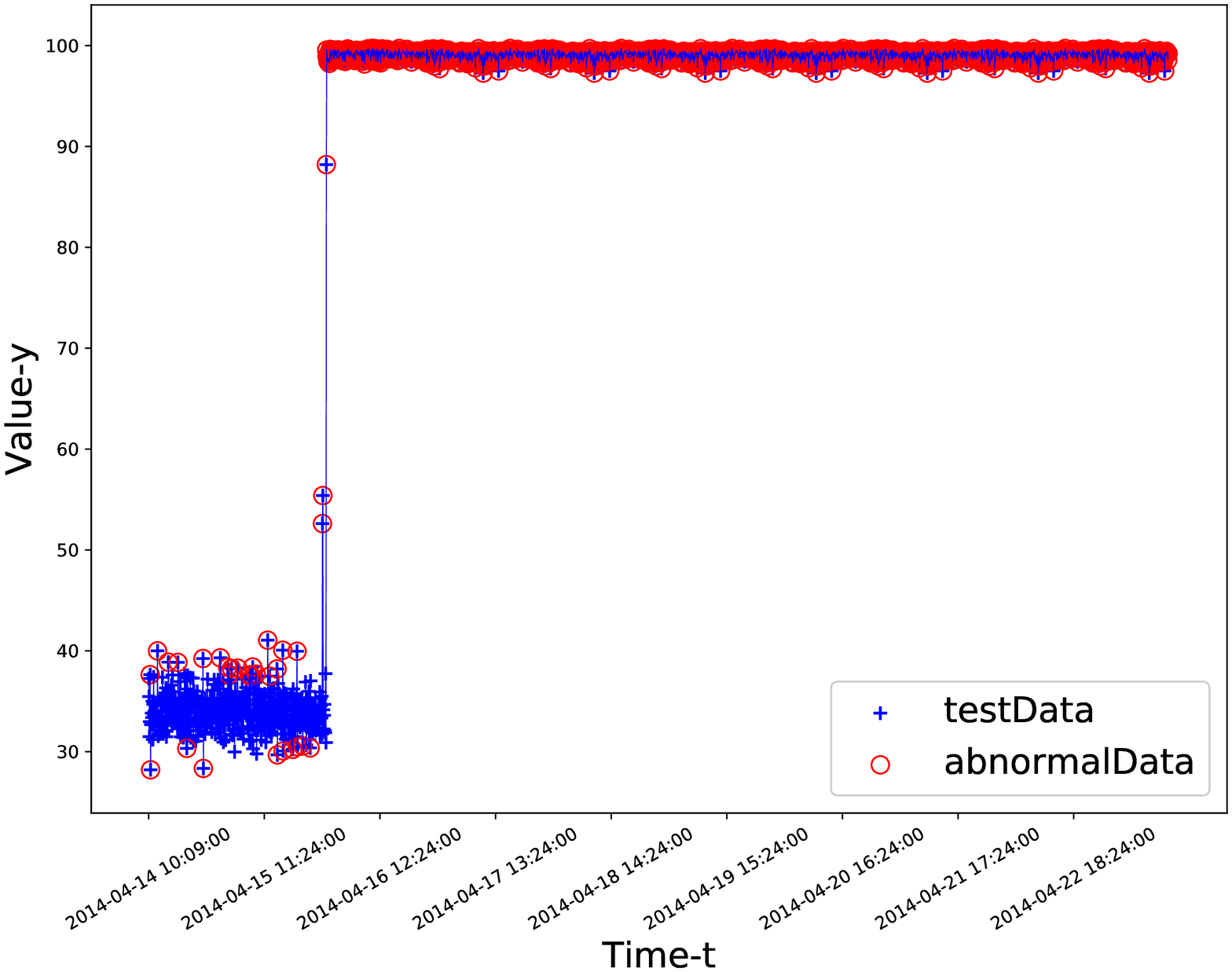}}
\subfigure[SGP-Q]
{\includegraphics[width=2.7in]{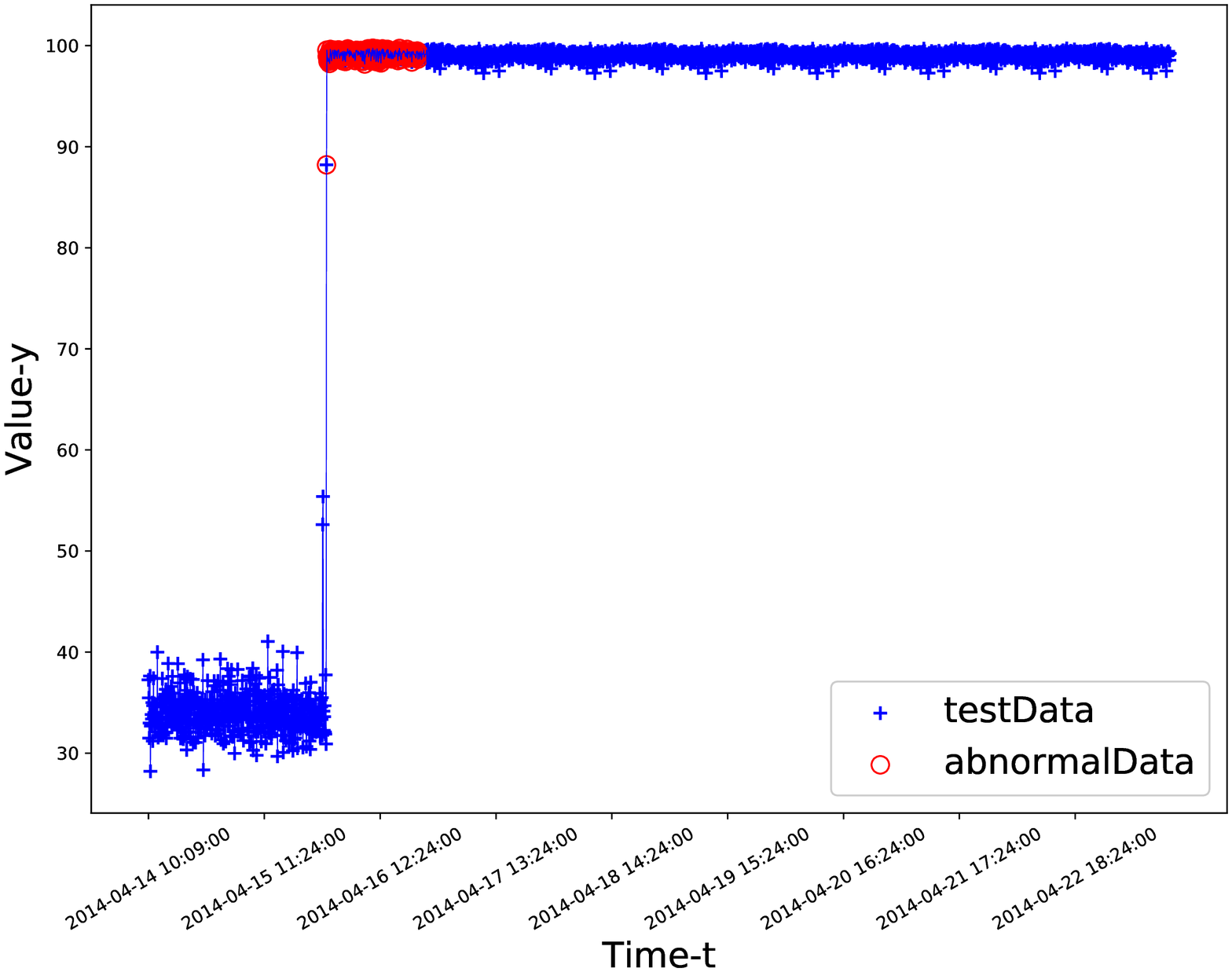}}
\subfigure[Label]
{\includegraphics[width=2.7in]{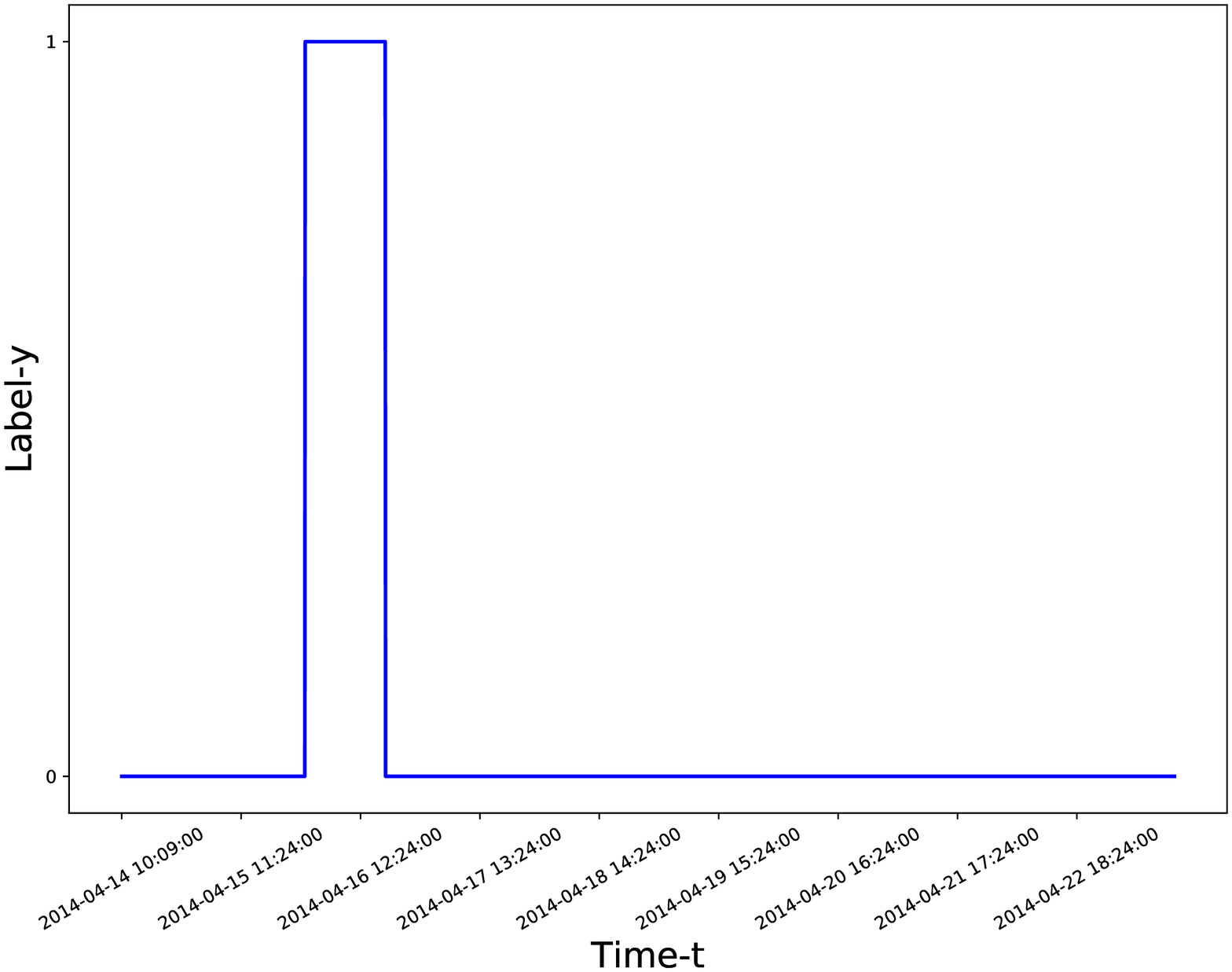}}
\caption{(a), (b), (c) and (d) show the results of anomaly detection of four methods on the `ec2\_cpu\_utilization\_ac20cd' dataset. (e) shows labels of test data, where 1 represents abnormal data and 0 represents normal data.}\label{fig:ec2cpuutilizationac20cd}
\vspace{-0.1in}
\end{figure}

Figure \ref{fig:ec2cpuutilization24ae8d}, Figure \ref{fig:ec2cpuutilization825cc2}, Figure \ref{fig:grokasganomaly} and Figure \ref{fig:ec2cpuutilizationac20cd} show the results of four methods on the datasets from the AWS server. Figure \ref{fig:ec2cpuutilization24ae8d} shows that the results of anomaly detection by the four methods are similar on the `ec2\_cpu\_utilization\_24ae8d' dataset. In the Figure \ref{fig:ec2cpuutilization825cc2}, there is a slight concept dript after 2014-01-16 23:04:00. Specifically, the data in the latter part change more widely and their values are smaller than that of the data in the former part.
As shown in Figure \ref{fig:ec2cpuutilization825cc2}, when data have a slight concept drift, the GPR-IADAM can add new data with small deviations from the prediction mean into the training data to update the model, and thus the GPR-IADAM can address the slight concept drift. The GPR-ADAM cannot handle any concept drift at all, so the GPR-ADAM has the worst performance. The GPR-AD always adds true data to the training data to update the model, so the GPR-AD can also deal with concept drift. The SGP-Q considers the abnormal degree of the current data point relative to that of previous data to decide whether to add the data point to train model, which is fully capable of handling concept drift. The proposed SGP-Q has the best performance on the `ec2\_cpu\_utilization\_825cc2' dataset.

In Figure \ref{fig:grokasganomaly} and Figure \ref{fig:ec2cpuutilizationac20cd}, concept drift occurs in the time-series data, and the deviation between new data and old data is huge. In this case, the data that initially change are considered as abnormal data. However, when the `abnormal' behaviors last for a while, the new data should be considered as normal data, and the model should relearn the characteristics of normal data. As shown in Figure \ref{fig:grokasganomaly} and Figure \ref{fig:ec2cpuutilizationac20cd}, the GPR-ADAM and GPR-IADAM  are unable to deal with the concept drift when new data deviate greatly from old data, and all data after the concept drift are marked as abnormal data. Therefore, the GPR-ADAM and GPR-IADAM perform poorly on the `grok\_asg\_anomaly' and `ec2\_cpu\_utilization\_ac20cd' datasets.
Whether data are abnormal or not, the GPR-AD always adds true data instead of prediction mean to update the GPR model.
Therefore, the GPR-AD can deal with concept drift in this case by adding new data to update the GPR model, and thus the performance of the GPR-AD is better than that of the GPR-ADAM and GPR-IADAM. The SGP-Q considers the information of previous and current data rather than the information of the current data point to measure the abnormal degree of the current data point, which is more reasonable. Therefore, the SGP-Q can deal with the concept drift well when the deviation between new data and old data is large. The SGP-Q has the best performance on the `grok\_asg\_anomaly' and `ec2\_cpu\_utilization\_ac20cd' datasets.

\begin{figure}[ht]
\centering
\subfigure[GPR-AD]
{\includegraphics[width=2.7in]{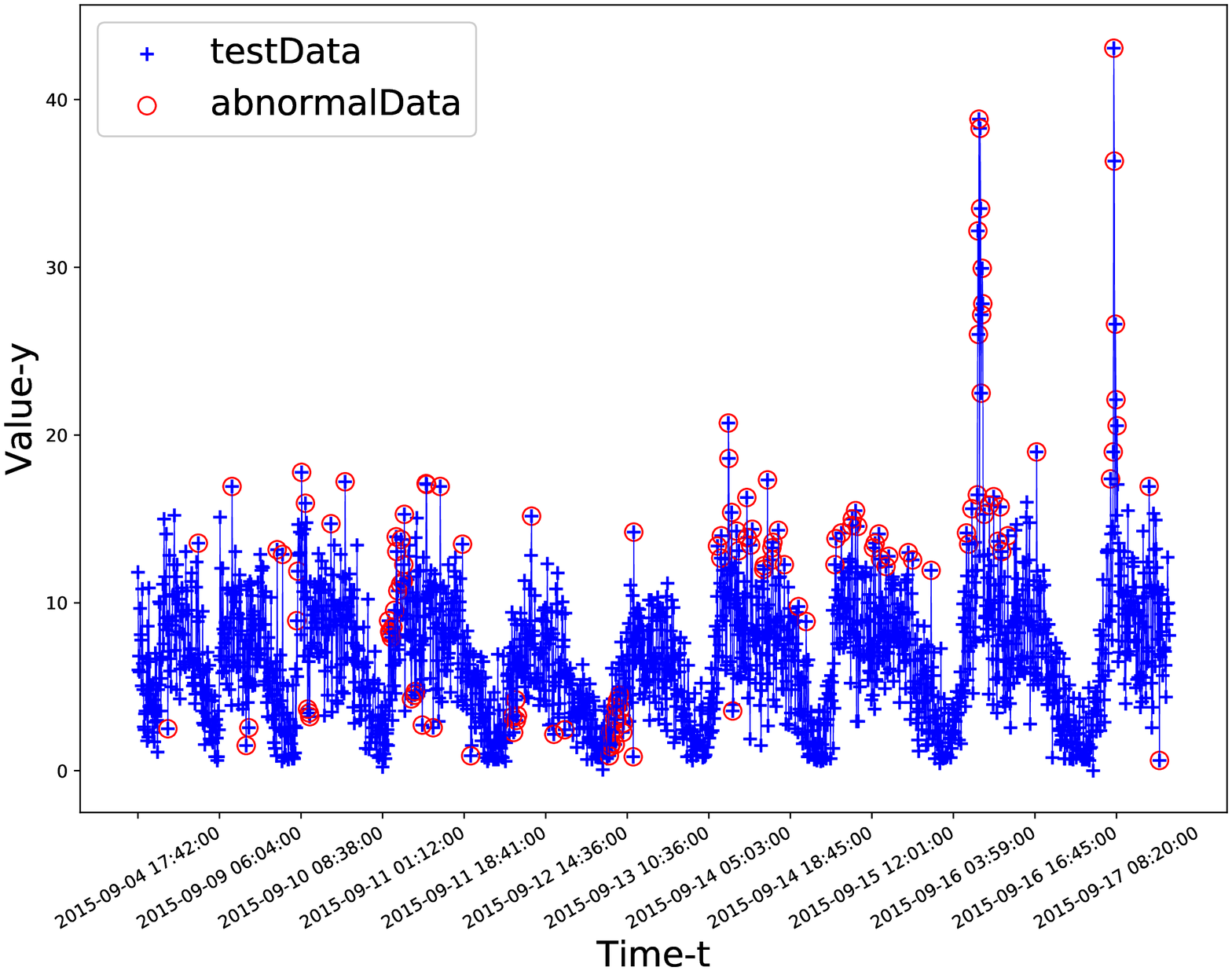}}
\subfigure[GPR-ADAM]
{\includegraphics[width=2.7in]{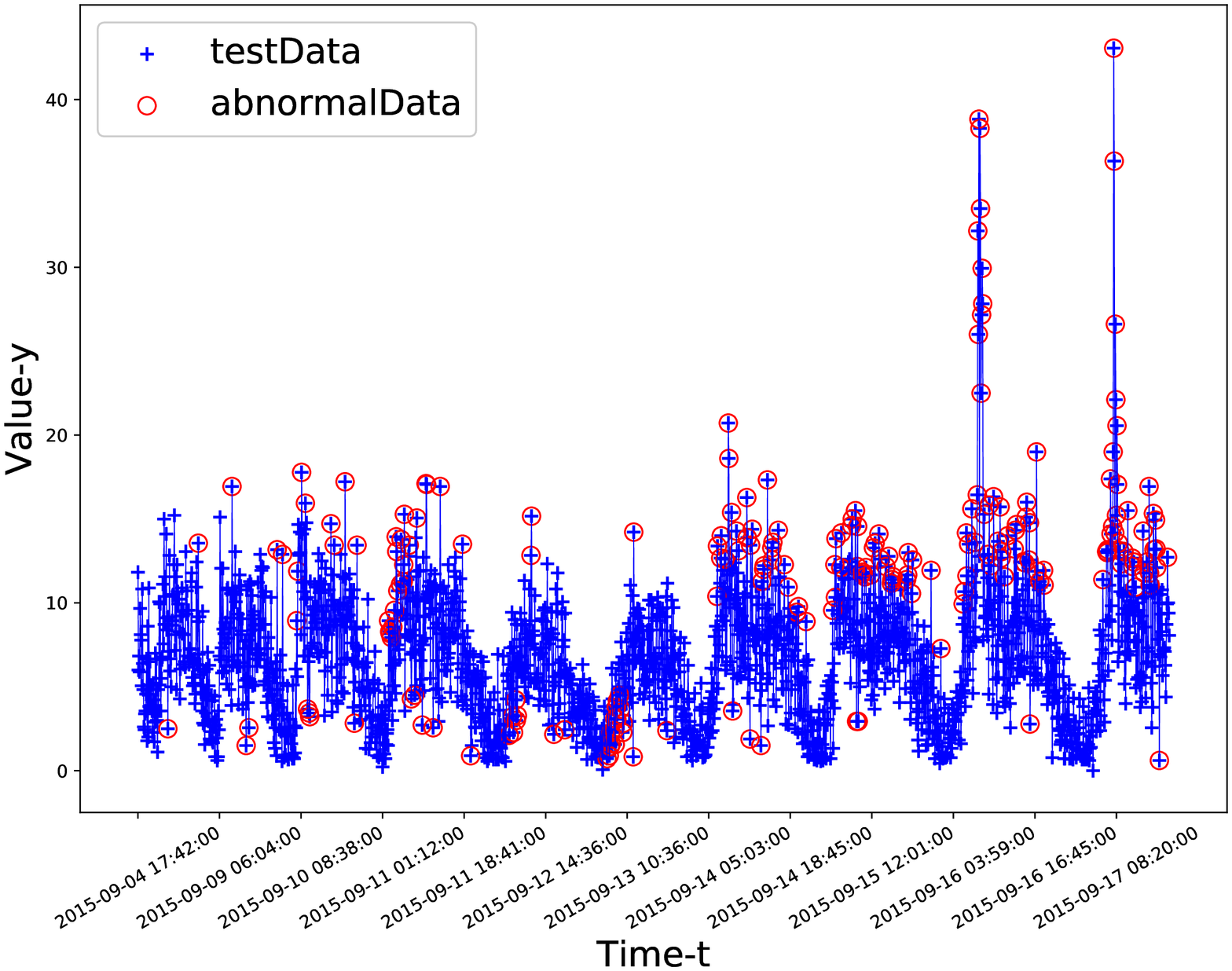}}
\subfigure[GPR-IADAM]
{\includegraphics[width=2.7in]{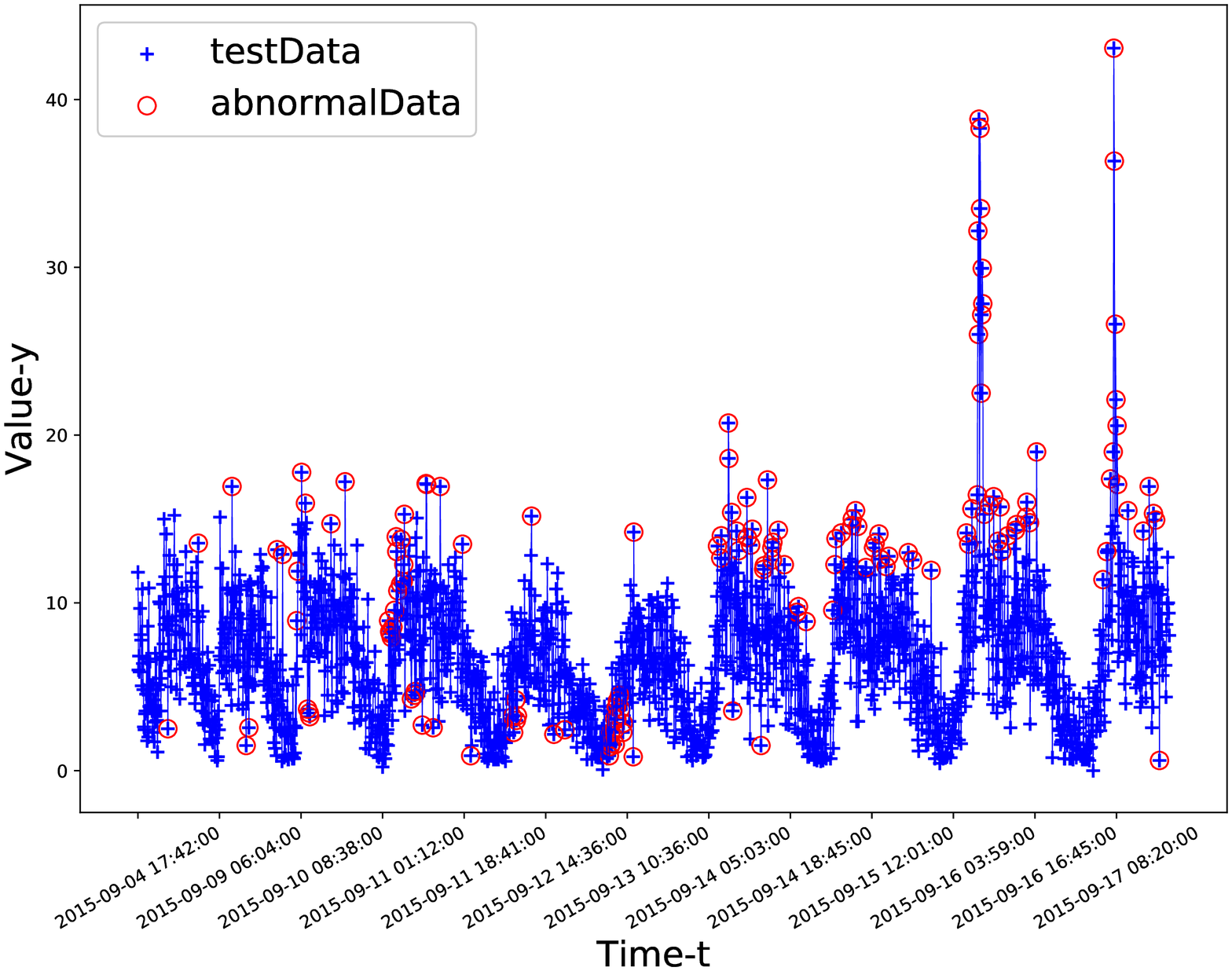}}
\subfigure[SGP-Q]
{\includegraphics[width=2.7in]{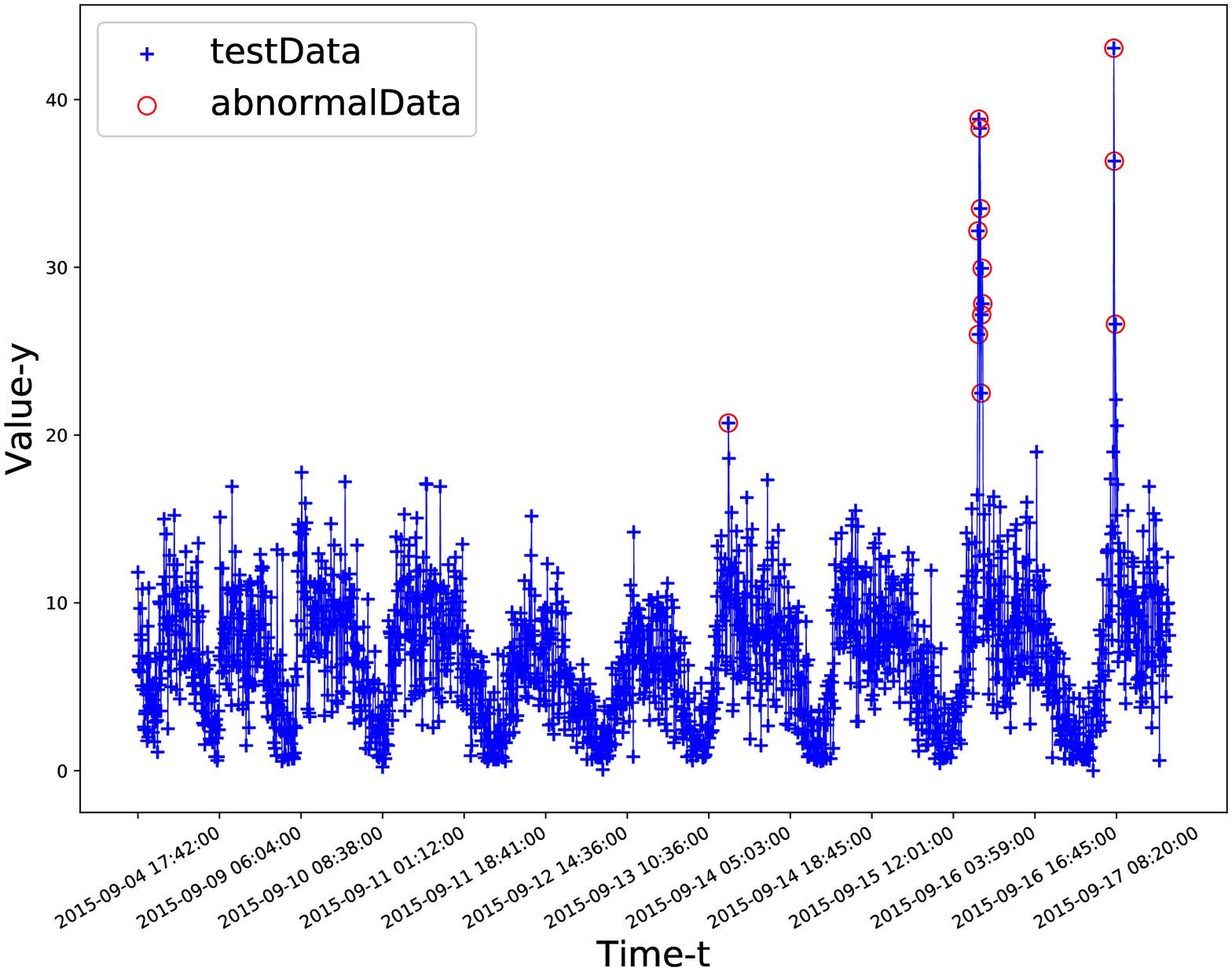}}
\subfigure[Label]
{\includegraphics[width=2.7in]{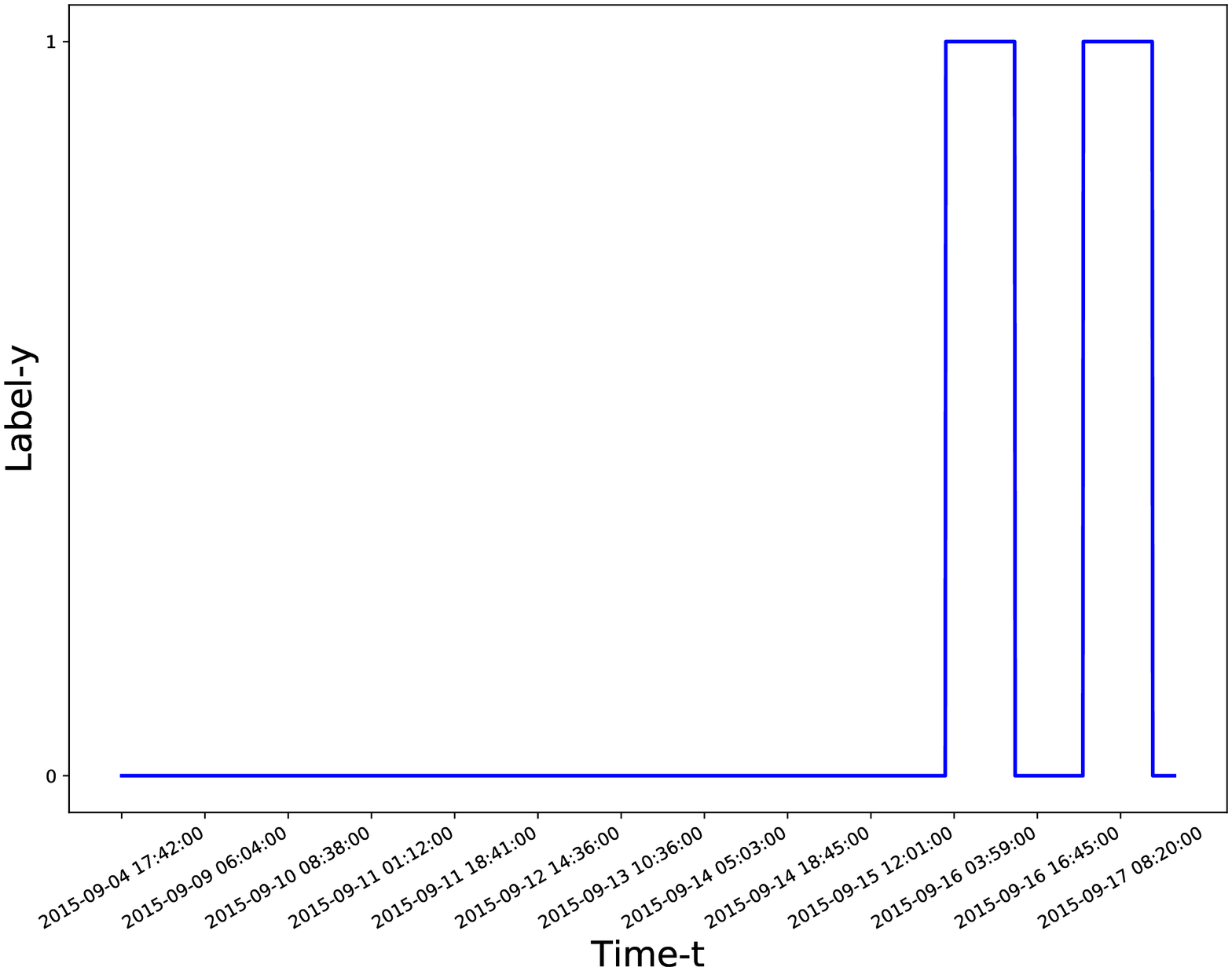}}
\caption{(a), (b), (c) and (d) show the results of anomaly detection of four methods on the `occupancy\_t4013' dataset. (e) shows labels of test data, where 1 represents abnormal data and 0 represents normal data.}\label{fig:occupancyt4013}
\vspace{-0.1in}
\end{figure}

\begin{figure}[ht]
\centering
\subfigure[GPR-AD]
{\includegraphics[width=2.7in]{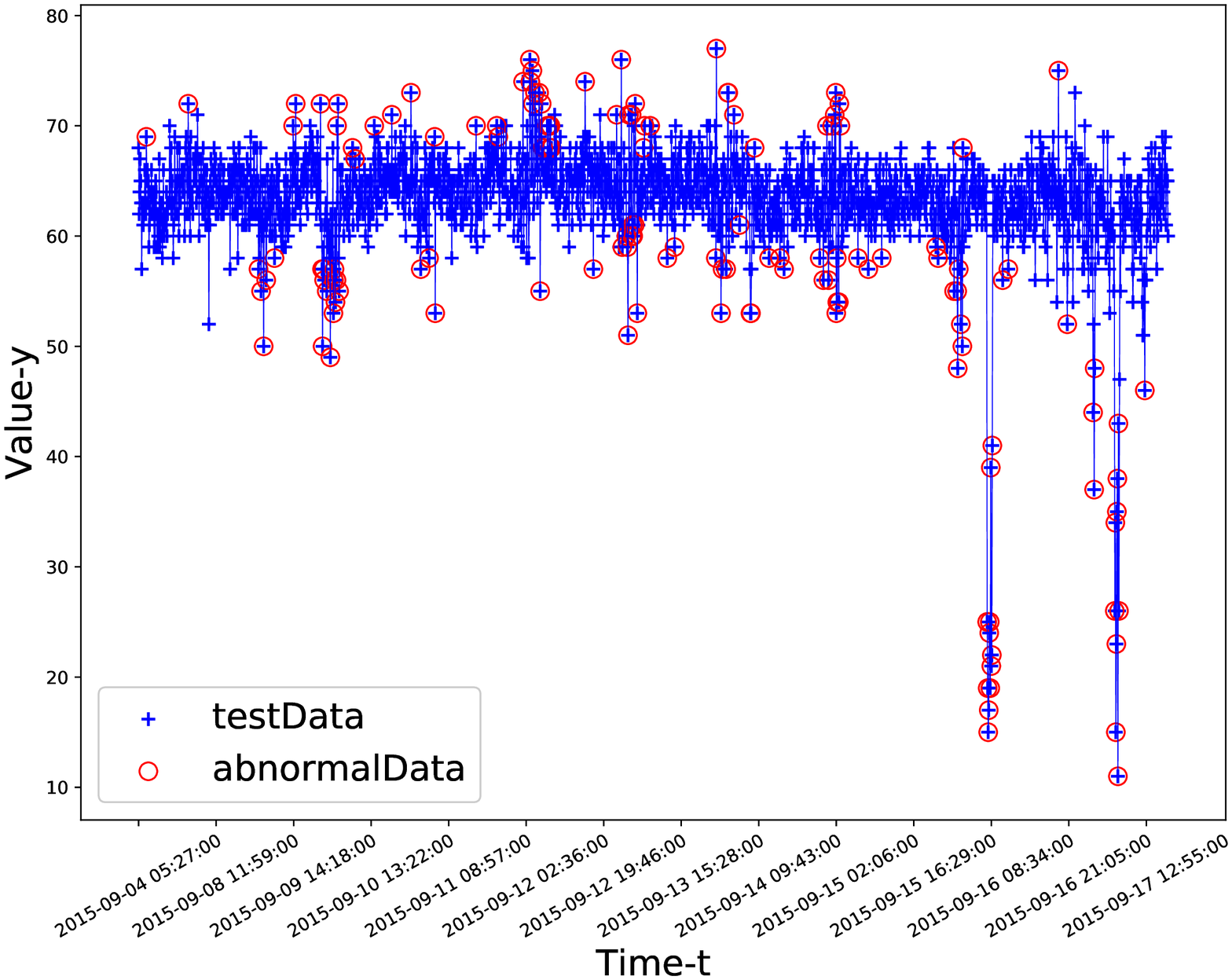}}
\subfigure[GPR-ADAM]
{\includegraphics[width=2.7in]{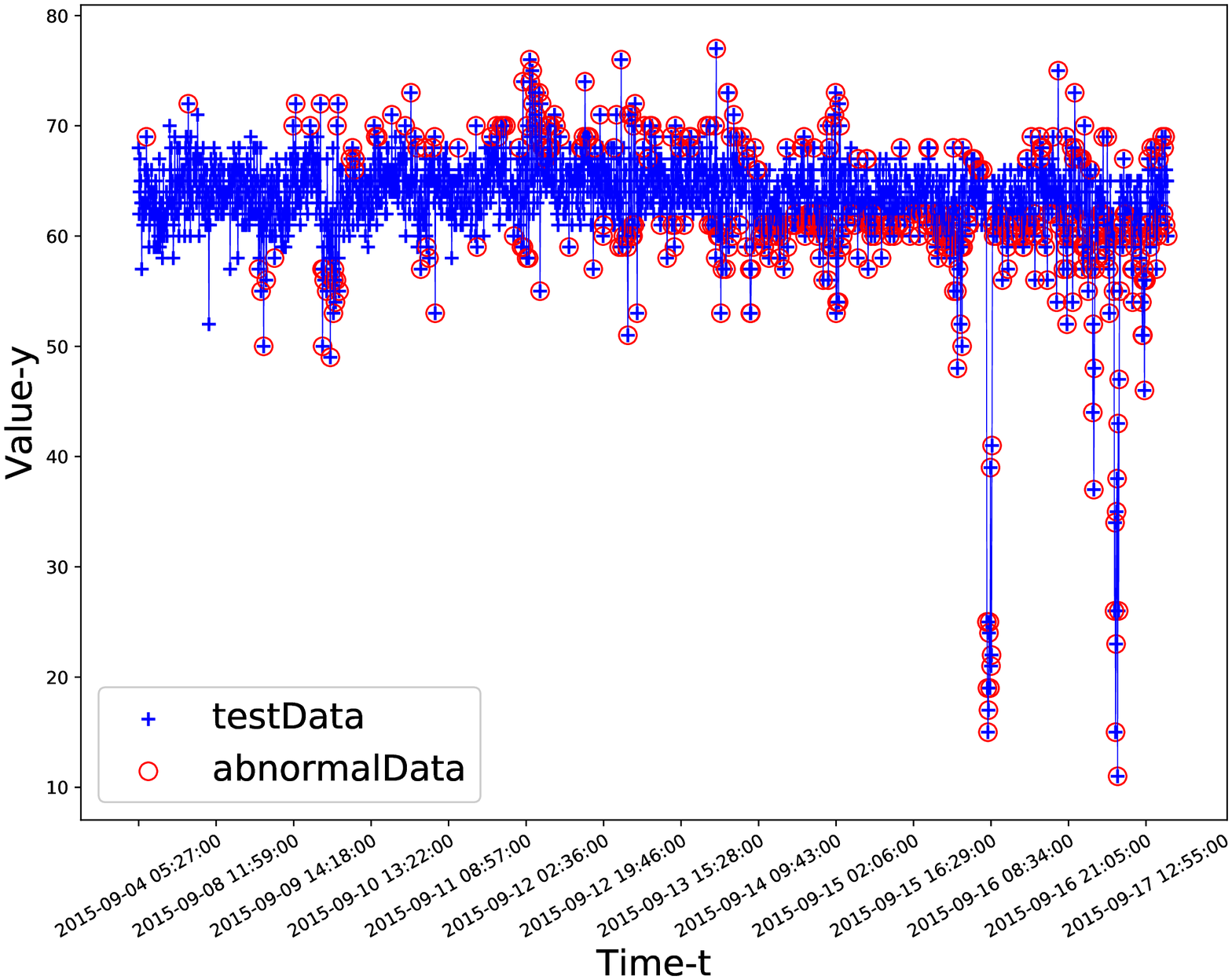}}
\subfigure[GPR-IADAM]
{\includegraphics[width=2.7in]{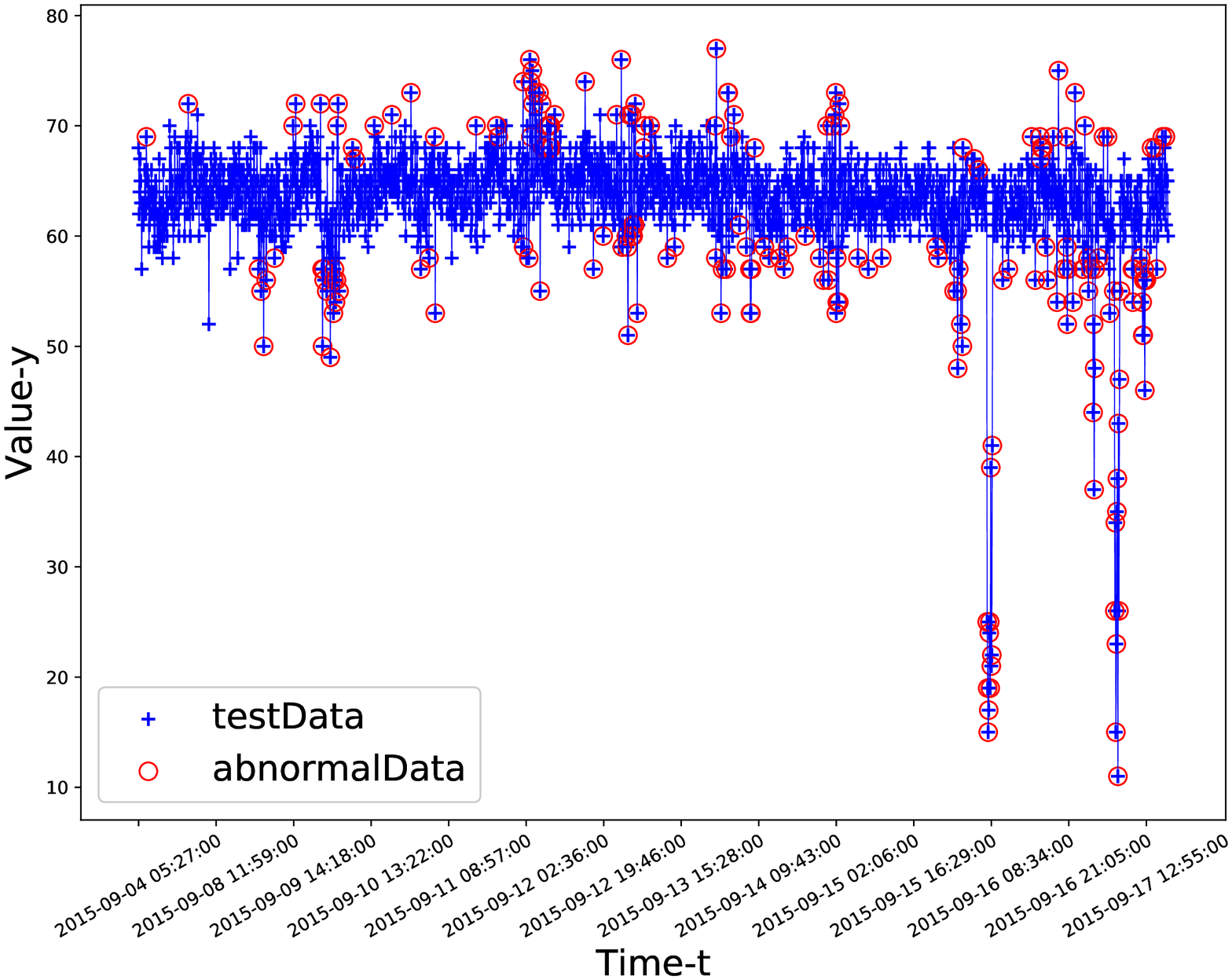}}
\subfigure[SGP-Q]
{\includegraphics[width=2.7in]{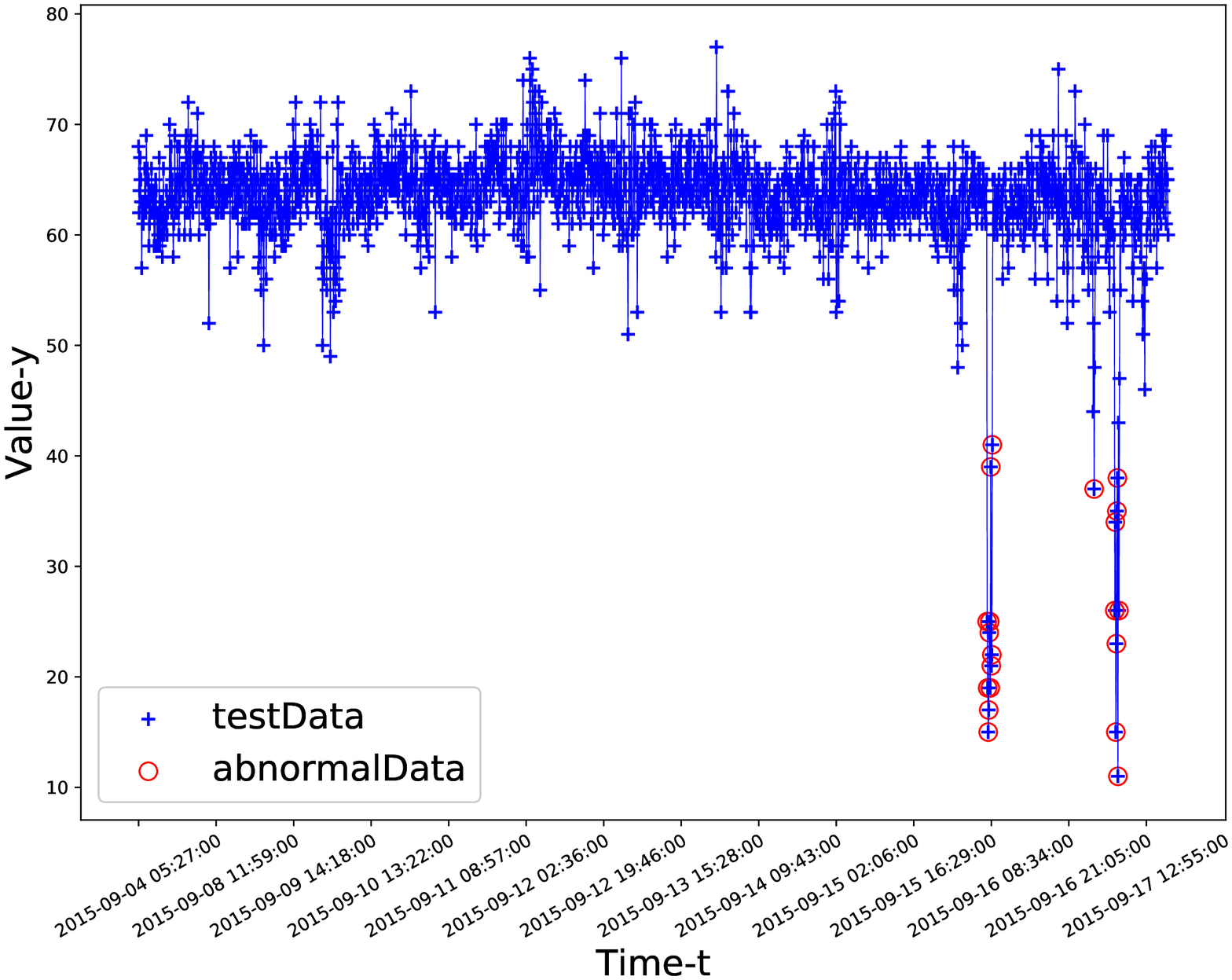}}
\subfigure[Label]
{\includegraphics[width=2.7in]{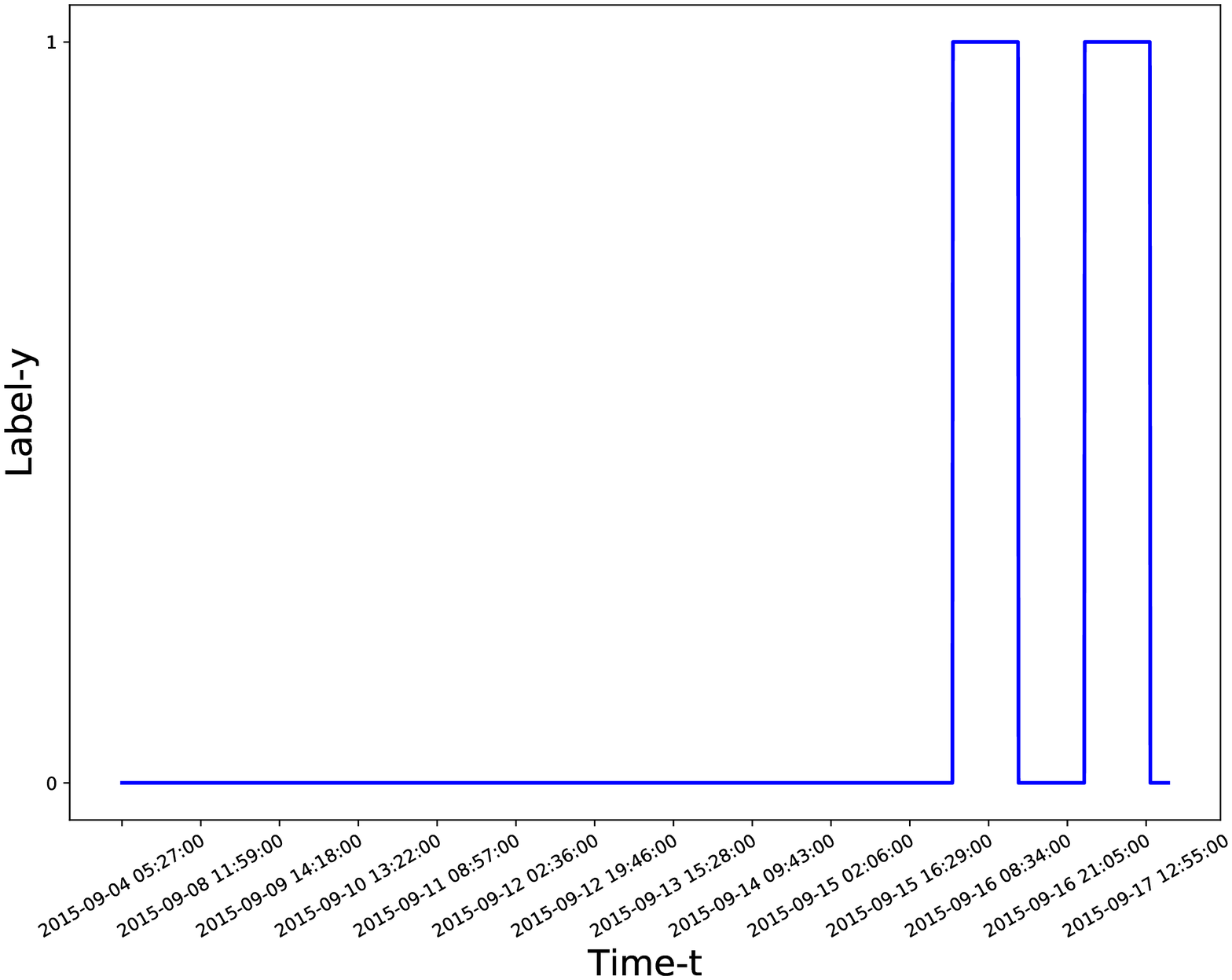}}
\caption{(a), (b), (c) and (d) show the results of anomaly detection of four methods on the `speed\_t4013' dataset. (e) shows labels of test data, where 1 represents abnormal data and 0 represents normal data.}\label{fig:speedt4013}
\vspace{-0.1in}
\end{figure}

Figure \ref{fig:occupancyt4013} and Figure \ref{fig:speedt4013} show results of four methods on the real-time traffic data. From Figure \ref{fig:occupancyt4013}
and \ref{fig:speedt4013}, we can see that the GPR-AD, GPR-ADAM and GPR-IADAM mark many normal data as abnormal data. The number of normal data that are marked as abnormal data is the lowest in the SGP-Q. Combined with the numerical results in Table \ref{F1Results}, the SGP-Q performs best on the `occupancy\_t4013' and `speed\_t4013' datasets.

Our method SGP-Q achieves better performance than the GPR-AD, GPR-ADAM, and GPR-IADAM for both numerical and visual results. The reasons for good performance are listed as follows. Firstly, compared with the GPR-AD, the SGP-Q can make use of as few abnormal data as possible in the training data by the strategy based on likelihood and Q-function, which can make the SGPR model more accurate and improve the performance of anomaly detection. Secondly, the SGP-Q can address concept drift in data, while GPR-ADAM cannot. Thirdly, the GPR-IADAM only uses the information of the current data point to measure the abnormal degree of the current data point. Only when the deviation between the current data point and the prediction mean is small, the data point can be added to the training data. Therefore, the GPR-IADAM can only deal with concept drift with a slight deviation between new data and old data. The SGP-Q takes into account the information of previous and current data to measure the abnormal degree of the current data point relative to that of the previous data, which is more reasonable than only using the information of the current data point. In addition, whether the deviation between the new data and old data in the concept drift is large or slight, the SGP-Q can handle the concept drift well using Q-function. The SGP-Q overcomes the limitations of the GPR-AD, GPR-ADAM, and GPR-IADAM. Therefore, the performance of the SGP-Q is better than that of the GPR-AD, GPR-ADAM, and GPR-IADAM.


\subsection{Summary}
We summarize the applicable scenarios for each method.
Firstly, the GPR-AD updates the model with true data regardless of whether data are abnormal or not. On the one hand, the GPR-AD is able to handle concept drift. On the other hand, there are too many abnormal data in the training data, resulting in the inaccurate GPR models and wrong anomaly detection results.
Secondly, when the current data point is abnormal, the GPR-ADAM adds the prediction mean to update the GPR model and cannot address the concept drift. Thirdly, the GPR-IADAM uses the value of $\beta$ to measure the abnormal degree of the current data point. When the abnormal degree of the current data point is low, the data point is used to update the model; when the abnormal degree of the current data point is high, the prediction mean is used to update the model. Therefore, the GPR-IADAM can only deal with the concept drift when the deviation between new data and old data is slight. Fourthly, the SGP-Q uses the information of previous and current data instead of the information of the current data point to measure the abnormal degree of the current data point relative to that of previous data. Regardless of whether the deviation between new data and old data is large or slight, the SGP-Q can address the concept drift well. In addition, the SGP-Q can obtain more accurate SGPR models and better anomaly detection results by making use of few abnormal data in the training data through the strategy based on likelihood and Q-function.

The number of points in the inducing input $Z$ is set to 100. Increasing the number of points in the inducing input will increase the training time, while the performance improvement is tiny. The covariance function in the experiment is set to the sum of the RBF kernel function and the linear kernel function. We have tried some more complex kernel functions, such as the multi-layer perceptron (MLP) kernel function and Matern32 kernel function. Complex kernel functions increase the number of kernel parameters that need to be optimized, resulting in much slower training speed but weeny performance improvement.

The actual time-series data are long sequences with timestamps. For the convenience of calculation, the timestamps must be quantized. Two numerical methods have been tried. The first method is that the time of the first data point in the time-series data is recorded as 0, and the subsequent time is quantized to the number of minutes between the current time and start time multiplied by 0.01. The second method is to slice long time-series data, the daily 00:00:00 is quantized to 0, and the other time is quantized to the number of minutes between the current time and today's 00:00:00 multiplied by 0.01. In the first numerical method, modeling the mapping from input to output requires more complex composite kernel functions. For example, modeling periodic data needs to add a periodic kernel function to the composite kernel function. In the second numerical method, a relatively simple composite kernel function, that is, the sum of the RBF kernel function and the linear kernel function is able to model the mapping from input to output well. Even if the data is periodic, there is no need to add the periodic kernel function into composite kernel function, because the periodic information is already included in the numerical time.
The second numerical method combined with a relatively simple kernel function is faster than the first numerical method combined with a complex kernel function, and their performance is quite similar. In the experiment, we use the second numerical method, and the experimental results confirm the validity and rationality of the second numerical method.

\section{Conclusion}
In this paper, we have proposed the SGP-Q method, which improves the existing online anomaly detection methods based on GPs. As an online anomaly detection method, the SGP-Q uses SGPs with lower time complexity to model time-series data and accelerates online anomaly detection. Concept drift is common in time-series data, and it is essential for online anomaly detection methods to have the abilities to redefine the meanings of `abnormal' behaviors and adapt to concept drift. On account of using Q-function, the SGP-Q can address concept drift well. Moreover, the SGP-Q employs the strategy based on likelihood and Q-function to update training data. This strategy can reduce the abnormal data in the training data and make the SGPR model more accurate, thus improving the performance of anomaly detection.

In experiments, we conducted experiments on various artificial and real-world datasets and compared the proposed SGP-Q with the existing anomaly detection methods based on GPs, including the GPR-AD, GPR-ADAM, and GPR-IADAM. The proposed SGP-Q obtains better performance than the GPR-AD, GPR-ADAM, and GPR-IADAM.

For future work, it will be more challenge and interesting to employ the mixture of Gaussian processes to model time-series data in the task of online anomaly detection, as real-world time-series data are usually in the `multi-modality' distributions.

\section*{Acknowledgements}
This work is supported by the National Natural Science Foundation of China under Project 61673179.

\vskip 0.2in
\bibliographystyle{unsrt}
\bibliography{SGPQ}
\end{document}